\documentclass{article}

\usepackage[main,final]{neurips_2026}

\usepackage[utf8]{inputenc}
\usepackage[T1]{fontenc}
\usepackage{hyperref}
\usepackage{url}
\usepackage{booktabs}
\usepackage{amsfonts}
\usepackage{amsmath}
\usepackage{amssymb}
\usepackage{nicefrac}
\usepackage{microtype}
\usepackage{xcolor}
\usepackage{graphicx}
\usepackage{subcaption}
\usepackage{tikz}
\usepackage{wrapfig}
\usepackage{titletoc}

\newcommand{\R}{\mathbb{R}}

\newcommand{\rev}[1]{#1}
\newenvironment{revblock}{}{}

\title{Neural Harmonic Measure Operator}

\author{%
  Jinjin He \quad Sinan Wang \quad Yuchen Sun \quad Bo Zhu\\
  Georgia Institute of Technology\\
  \texttt{\{jhe433,swang3081,ysun748,bo.zhu\}@gatech.edu}
}

\begin{document}

\maketitle

\begin{abstract}
We introduce \textbf{Neural Harmonic Measure Operator (NHMO)}, a neural solver for elliptic PDE problems on variable-shape domains. The harmonic measure of a domain is the boundary probability distribution that, integrated against any boundary data, returns the Dirichlet Laplace solution. It depends only on the geometry, not on the boundary data. NHMO parameterizes \rev{the density of} this measure as a transformer-based boundary kernel supervised by Walk-on-Spheres exit samples, so one trained kernel handles different boundary \rev{values} on a shape with no retraining. We extend it to Poisson via a classical decomposition, with an auxiliary network amortizing the source-induced correction and avoiding the singular volume quadrature that breaks direct evaluation. At inference, new boundary values and new sources both yield PDE solutions by re-integration against the fitted kernel and lift, with no retraining. NHMO improves over four prior baselines on the MCB-B 3D variable-shape Poisson benchmark across all five categories, and is competitive with major neural-operator baselines on a controlled 2D testbed.
\end{abstract}

\section{Introduction}\label{sec:intro}

Classical solvers like the finite element method and finite differences~\citep{hughes2003finite,leveque2007finite} discretize the volumetric interior of $\Omega$ into a mesh and re-run the discretize-and-solve pipeline whenever the geometry, source, or boundary data changes, a bottleneck in design optimization~\citep{bendsoe2013topology}, uncertainty quantification~\citep{smith2024uncertainty}, and inverse problems~\citep{engl1996regularization}. Neural operators amortize this cost by learning a function-to-function map
$(\Omega, f, h) \mapsto u$ from domain, source, and boundary data to the solution, which evaluates in a single forward pass once trained. Foundational architectures parameterize integral kernels in the Fourier domain on regular grids~\citep{li2020fourier} or use a branch--trunk decomposition at fixed sample points~\citep{lu2021learning}; recent transformer-based variants handle irregular meshes via attention over mesh points or learned slice tokens~\rev{\citep{wu2024transolver,wang2024latent,alkin2024universal,zhou2026transolver}}. Yet these methods inherit the volumetric framing of FEM/FDM, still operating on the bulk interior of $\Omega$ with compute scaling with volumetric discretization rather than the codimension-one boundary.
\begin{figure}
\centering
\includegraphics[width=0.95\linewidth]{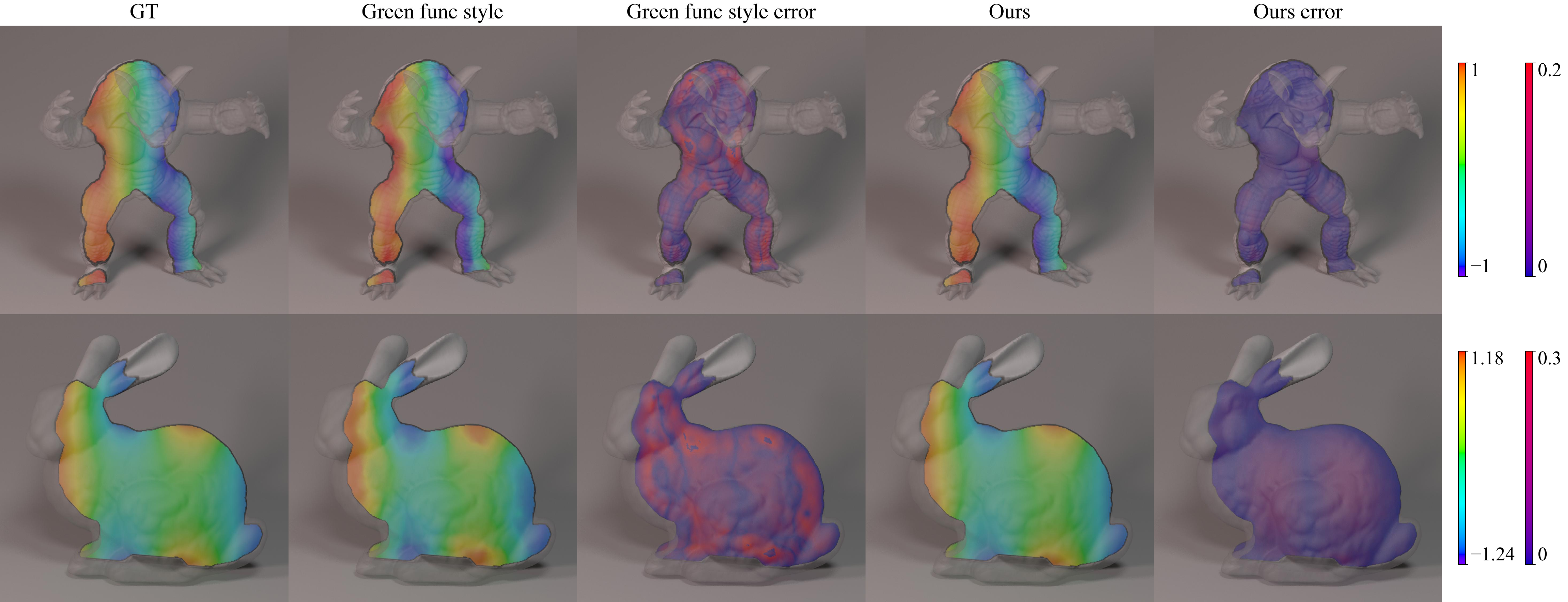}
\caption{Intuitive demonstration on complex 3D shapes (top: armadillo; bottom: bunny), with a single shape-conditioned $K_\theta$ encoding both. Mean rel-$L_2$ is $0.012$ vs $0.142$ (Ours vs GF style; \S\ref{sec:exp-intuitive}). Columns: GT \rev{(reference solution)}, \rev{the} Green's-function-style \rev{(GF-style)} baseline, its absolute error, our prediction, \rev{and} our absolute error. \rev{Per row, fields share the left color bar and errors the right.}}
\label{fig:teaser}
\end{figure}

Boundary integral methods take a different angle on the same problem. The Boundary Element Method (BEM) reformulates a volumetric Dirichlet problem as an integral equation against a Green's-function kernel on the boundary $\partial\Omega$ alone~\citep{sauter2010boundary}, but it still requires a discretization of $\partial\Omega$ and dense linear-system solves. Stochastic methods such as Walk-on-Spheres (WoS)~\citep{muller1956some,sawhney2020monte,sawhney2022grid,sawhney2023walk} sidestep boundary discretization by estimating $u(p) = \mathbb{E}_p[h(B_\tau)]$ via Brownian-exit Monte Carlo, where $\mathbb{E}_p$ is the expectation over Brownian motions $B_t$ started at $p$ and $B_\tau$ is the first-exit point on $\partial\Omega$, but remains a per-query estimator instead of an amortized operator, paying $O(N_{\text{walks}})$ each time the boundary datum changes. Recent learned Green's-function-style operators, including NGF~\citep{yoo2025neural} and others~\citep{gin2021deepgreen,li2020multipole,teixeira2026variational}, all parameterize a volumetric kernel on the full pair space $\Omega \times \Omega$ and inherit the singular Green's function (see \S\ref{sec:related}).

We propose \textbf{Neural Harmonic Measure Operator (NHMO)}, a boundary-only neural operator for elliptic PDEs on variable-shape domains that, in contrast to the volumetric Green's-function operators above, learns \rev{the density of} a codimension-one boundary measure on $\Omega \times \partial\Omega$ rather than a kernel on $\Omega \times \Omega$. This drops the kernel domain by one dimension and replaces a singular volumetric kernel with a \rev{probability density on the boundary}. NHMO contains two learned components. First, a transformer-based boundary kernel $K_\theta(p, \zeta; \Omega)$ approximates \rev{the density $d\omega_p/d\sigma$ of} the harmonic measure $\omega_p$, a geometry-only distribution over the boundary\rev{; we call $K_\theta$ the harmonic-measure density}. Here, geometry-only means that $\omega_p$ depends on the domain $\Omega$ and query point $p$, but not on the prescribed boundary values. Integrating this distribution against any boundary data then recovers the Dirichlet Laplace solution via Kakutani's representation~\citep{kakutani1944143}. Second, a residual lift $v_\varphi$ \rev{carries} the source-induced contribution for Poisson problems via the classical balayage decomposition. The two components compose additively. These give NHMO structural advantages over volumetric operators. As a geometry-only probability kernel that depends on $\Omega$ rather than $h$ or $f$, a single fitted $K_\theta$ can be reused for arbitrary boundary data on the same shape without retraining, \rev{and, once the kernel is normalized over the boundary quadrature, its boundary term satisfies} the maximum principle by construction. It is trained mesh-free from Walk-on-Spheres exit samples, requiring no FEM solutions or tetrahedral meshes for kernel supervision. As a proof of concept, Figure~\ref{fig:teaser} shows NHMO and a Green's-function-style baseline on two complex 3D shapes (detailed in \S\ref{sec:exp-intuitive}).



\paragraph{Contributions.} \textbf{(1)} We propose to encode the \rev{density of the} harmonic measure $\omega_p$ as a learnable boundary kernel $K_\theta$ that depends on the geometry $\Omega$ alone (independent of boundary data $h$ and source $f$), supervised by Walk-on-Spheres (\S\ref{sec:method-kernel}, \S\ref{sec:exp-kernel-quality}). \textbf{(2)} We extend NHMO to Poisson problems via the classical balayage decomposition, with a zero-boundary-gauge lift $v_\varphi$ that reuses $K_\theta$ to amortize the source correction without singular volume quadrature (\S\ref{sec:method-lift}). \textbf{(3)} We validate NHMO on the MCB-B 3D Poisson benchmark\rev{, where it outperforms four neural-operator baselines, and on} a controlled 2D MNIST testbed, and show the framework's generality through intuitive 3D harmonic and drift-adaptation experiments (\S\ref{sec:exp-mcb}, \S\ref{sec:exp-mnist}, \S\ref{sec:exp-intuitive}).

\section{Related Work}\label{sec:related}

\paragraph{Neural operators for PDEs.} Function-to-function neural operators~\citep{kovachki2023neural} learn end-to-end maps from problem data to solutions. Foundational architectures include Graph Neural Operators~\citep{li2020neural}, Fourier Neural Operators~\citep{li2020fourier} that parameterize integral kernels in the spectral domain, and DeepONet~\citep{lu2021learning} with a branch-trunk architecture on functions sampled at fixed points. Geometry-aware extensions handle irregular meshes via attention or graph modules~\rev{\citep{li2023fourier,wu2024transolver,wang2024latent,alkin2024universal}}, and several lines target varying domain geometries~\citep{wang2024beno,yin2024dimon,wu2026geopt}. Transformer-based operators~\citep{hao2023gnot,xiao2023improved,luo2025transolver++,zhou2026transolver} use attention over mesh points or learned slice tokens. These methods regress the solution or solution operator directly; we instead model the boundary measure that mediates all solutions.

\paragraph{Learning Green's functions and integral operators.} A separate line learns the volumetric Green's function $G_\Omega(p,q)$ for linear PDEs via rational neural networks~\citep{boulle2022data}, Dirac-delta approximations~\citep{teng2022learning}, radial-basis approximations~\citep{negi2024learning}, and variational principles~\citep{teixeira2026variational}; deep nonlinear-BVP extensions appear in DeepGreen~\citep{gin2021deepgreen}, and Green's-function-style multipole structure underlies the multipole graph neural operator~\citep{li2020multipole}. Neural Green's Functions (NGF)~\citep{yoo2025neural} is the closest prior work and our principal baseline. NGF learns the domain Green's function $G_\Omega(p,q)$ as $\Phi_\theta(p)^\top D \Phi_\theta(q)$ for learned per-point features, \rev{trained on precomputed FEM solution fields,} and recovers solutions by integrating $f$ against $G_\Omega$ in the volume and $h$ against the outward-normal derivative on the boundary. Earlier 2D boundary-integral neural methods~\citep{lin2021binet,sun2023binn} and neural integral operators~\citep{zappala2024learning} target classical BEM-style discretizations rather than amortizing across boundary measures of varying shapes.

\paragraph{Walk on Spheres and grid-free Monte Carlo solvers.} Walk on Spheres (WoS)~\citep{muller1956some} is a Monte Carlo estimator for elliptic PDEs based on Brownian-exit simulation. The grid-free perspective was revived for graphics and learning by~\citet{sawhney2020monte}, and Walk on Stars~\citep{sawhney2023walk} extends it to mixed boundary conditions and source terms, with follow-ups for spatially varying coefficients, surface PDEs, gradient computation, and variance reduction~\citep{sawhney2022grid,sugimoto2024projected,miller2024differential,huang2025guiding,Zombie}. \rev{The ideal WoS estimator is unbiased; practical walks stop in an $\varepsilon$-shell around $\partial\Omega$ after $O(\log(1/\varepsilon))$ expected steps~\citep{binder2012rate}, introducing an $O(\varepsilon)$ bias~\citep{mascagni2003epsilon}. WoS} pays $O(N_{\text{walks}})$ per query at inference; neural surrogates trained against WoS targets~\citep{nam2024solving,zhang2025monte,miller2023boundary} amortize this cost. We use WoS as ground-truth supervision for $K_\theta$ rather than as a runtime estimator.

\paragraph{Harmonic measure in analysis.} The harmonic measure is classical in potential theory and geometric function theory~\citep{garnett2005harmonic}. In 2D it is conformally invariant, and its dimensional properties characterize boundary regularity~\citep{makarov1985distortion,armitage2012classical}. Kakutani's theorem~\citep{kakutani1944143} identifies it with the Brownian-exit law. To our knowledge, this is the first work to parameterize the \rev{density of the} harmonic measure with a neural network and to realize the balayage decomposition with a learned source amortizer.

\section{Background}\label{sec:background}

\subsection{Harmonic measure}\label{sec:bg-harmonic}

\begin{wrapfigure}{r}{0.32\linewidth}
\centering
\vspace{-1.0em}
\includegraphics[width=0.99\linewidth]{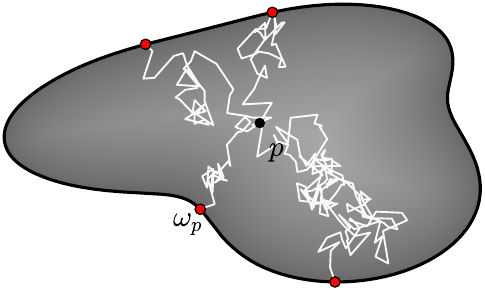}
\caption{Harmonic measure $(\omega_p)$ from Brownian exit locations. Red dots denote small boundary patches (E).}
\label{fig:blob-brownian}
\vspace{-0.5em}
\end{wrapfigure}
Let $\Omega \subset \R^d$ ($d \in \{2,3\}$) be a bounded Lipschitz domain. The \emph{harmonic measure} $\omega_p$ at $p \in \Omega$ is the probability distribution on $\partial\Omega$ describing where a Brownian motion started at $p$ first exits $\Omega$~\citep{kakutani1944143}: with $B_t$ a Brownian motion in $\R^d$ with $B_0 = p$ and $\tau = \inf\{t > 0 : B_t \notin \Omega\}$ its first-exit time,
\begin{equation}\label{eq:kakutani}
    \omega_p(E) \;=\; \mathbb{P}_p\!\left[B_\tau \in E\right], \qquad E \subset \partial\Omega \text{ Borel}.
\end{equation}
The family $\{\omega_p\}_{p \in \Omega}$ depends only on the geometry $\Omega$, not on any boundary data.

\paragraph{Constructing the Dirichlet solution.} For continuous boundary data $h \in C(\partial\Omega)$, the Dirichlet problem $\Delta u = 0$ in $\Omega$ with $u = h$ on $\partial\Omega$ has the closed-form solution~\citep{garnett2005harmonic}
\begin{equation}\label{eq:poisson-rep}
    u(p) \;=\; \int_{\partial\Omega} h(\zeta)\, d\omega_p(\zeta) \;=\; \mathbb{E}_p\!\left[h(B_\tau)\right].
\end{equation}
The probabilistic form on the right is the basis of Walk-on-Spheres Monte Carlo solvers~\citep{muller1956some,sawhney2020monte,sawhney2023walk}. Once $\omega_p$ is known for a geometry $\Omega$, equation~\eqref{eq:poisson-rep} resolves the Dirichlet problem for any $h$ via a single boundary integral. \rev{On a Lipschitz domain, $\omega_p$ is absolutely continuous with respect to the surface measure $\sigma$, and its Radon--Nikodym density $d\omega_p/d\sigma(\zeta) = -\partial_{\nu_\zeta} G_\Omega(p,\zeta)$ is the Poisson kernel, with $G_\Omega$ the Dirichlet Green's function. NHMO learns this \emph{harmonic-measure density}; we reserve ``harmonic measure'' for $\omega_p$ itself and name the method after it, since $\omega_p$ exists on any bounded domain and WoS exit points are drawn from it.} We present a derivation of~\eqref{eq:poisson-rep} and further properties of $\omega_p$ in Appendix~\ref{app:harmonic-measure}.

\subsection{Newtonian potential and balayage}\label{sec:bg-balayage}

The Poisson problem extends~\eqref{eq:poisson-rep} to nonzero source $f \in L^\infty(\Omega)$:
\begin{equation}\label{eq:poisson-bvp}
    \Delta u \;=\; f \text{ in } \Omega, \qquad u \;=\; h \text{ on } \partial\Omega.
\end{equation}
Let $\Phi$ be the fundamental solution of $-\Delta$ on $\R^d$\rev{, the radial solution of $-\Delta \Phi = \delta_0$, which is positive near the origin (the usual sign convention in potential theory):}
\begin{equation}\label{eq:fundamental}
    \Phi(x) \;=\; \begin{cases} -\tfrac{1}{2\pi}\log|x| & d = 2,\\[2pt] \tfrac{1}{4\pi|x|} & d = 3, \end{cases}
\end{equation}
and define the \emph{Newtonian potential} of $f$ by $N_f(p) = -\int_\Omega \Phi(p-q)\, f(q)\, dq$, so that $\Delta N_f = f$ on $\R^d$.

\paragraph{Balayage decomposition.} Setting $w = u - N_f$ in~\eqref{eq:poisson-bvp} yields $\Delta w = 0$ in $\Omega$ with $w|_{\partial\Omega} = h - N_f|_{\partial\Omega}$. Applying~\eqref{eq:poisson-rep} to $w$ \rev{and grouping the terms that do not involve $h$ gives}
\begin{revblock}
\begin{equation}\label{eq:balayage}
    u(p) \;=\; \int_{\partial\Omega} h(\zeta)\, d\omega_p(\zeta) \;+\; u_f(p),
    \qquad
    u_f(p) \;=\; N_f(p) \;-\; \int_{\partial\Omega} N_f|_{\partial\Omega}(\zeta)\, d\omega_p(\zeta),
\end{equation}
\end{revblock}
\rev{where the source-only piece $u_f$ is independent of $h$ and satisfies $\Delta u_f = f$ in $\Omega$ with $u_f|_{\partial\Omega} = 0$.} The same harmonic measure that handles the boundary data also handles the source-induced boundary correction\rev{, applied to $N_f|_{\partial\Omega}$ instead of $h$}. With $f \equiv 0$ the decomposition recovers the pure Laplace identity~\eqref{eq:poisson-rep}. NHMO's two-component factorization in \S\ref{sec:method} is the neural counterpart of this split: the boundary integral becomes a learned kernel, the source-only piece becomes a learned residual field.

\section{Neural Harmonic Measures}\label{sec:method}

Throughout this section, $\omega_p$ denotes the harmonic measure and $K_\theta$ the \rev{learned harmonic-measure density that approximates $d\omega_p/d\sigma$}; $v_\varphi$ denotes the residual lift. The full symbol list is in Appendix~\ref{app:notation}. \rev{Three equations play distinct roles: the continuous identity~\eqref{eq:balayage}, the learned model~\eqref{eq:nhmo-decomp}, and its quadrature implementation~\eqref{eq:bdy-int}.}

\subsection{Problem setup}\label{sec:problem}

Each shape category is a distribution over bounded Lipschitz domains $\Omega \subset \R^d$ with $d \in \{2, 3\}$. A training set provides shapes drawn from this distribution; per shape, a reference solution $u_{\text{true}}$ for a parametric family of Poisson problems $\Delta u = f$, $u|_{\partial\Omega} = h$ is given at a discretization of $\Omega$. The reference solver and discretization are experimental choices (\S\ref{sec:experiments}). At test time we evaluate on held-out shapes and on held-out $(h, f)$ pairs, including problems with coefficients drawn from outside the training support to test generalization across the parametric BC distribution.

\subsection{Decomposition}\label{sec:method-decomp}

\rev{By the balayage identity~\eqref{eq:balayage}, the solution splits into} a clean $h$-only boundary integral plus an $h$-independent source-only piece that vanishes on $\partial\Omega$. We factorize NHMO along this split:
\begin{equation}
    u(p) \;=\; \underbrace{\langle h,\, K_\theta(p,\cdot;\Omega)\rangle_{\partial\Omega}}_{u_h(p)} \;+\; \underbrace{v_\varphi(p;\Omega,h,f)}_{\approx\, u_f(p)},
    \label{eq:nhmo-decomp}
\end{equation}
where $u_h(p)$ denotes the boundary-integral prediction and $u_f(p)$ the zero-boundary Poisson particular solution; $K_\theta(p,\zeta;\Omega)$ is a learned \rev{harmonic-measure density approximating $d\omega_p/d\sigma$} and $v_\varphi$ is a learned residual field. Setting $K_\theta = \rev{d\omega_p/d\sigma}$ and $v_\varphi = u_f$ recovers~\eqref{eq:balayage} identically. In practice we let $v_\varphi$ depend on $h$ as well as $f$, so the residual lift can absorb approximation error from imperfect kernel fits on top of carrying the source contribution\rev{; \S\ref{sec:discussion} examines this dependence and separates the two roles}. Two properties of this decomposition are critical to our results. First, $K_\theta$ does not depend on $h$ or $f$, so the boundary kernel is geometry-only and a single fitted $K_\theta$ handles every $(h, f)$ pair on a shape without retraining. Second, $u_h$ is a linear functional of the boundary data: a coefficient shift in $h$, including coefficients drawn from outside the training support, changes $u_h$ proportionally without altering the kernel itself. End-to-end operators that fit $u$ as a nonlinear map of $(h, f)$ do not enjoy this property; their solution can drift arbitrarily under a coefficient shift unseen at training time. Out-of-distribution generalization across the parametric BC family is therefore a structural property of \rev{NHMO's kernel channel}, not an emergent effect from fitting\rev{; the lift carries no such guarantee}. The residual lift $v_\varphi$ catches source-induced contributions and any remaining approximation error.

\subsection{Boundary kernel $K_\theta$}\label{sec:method-kernel}

\begin{wrapfigure}{r}{0.52\linewidth}
\centering
\vspace{-1.2em}
\includegraphics[width=0.99\linewidth]{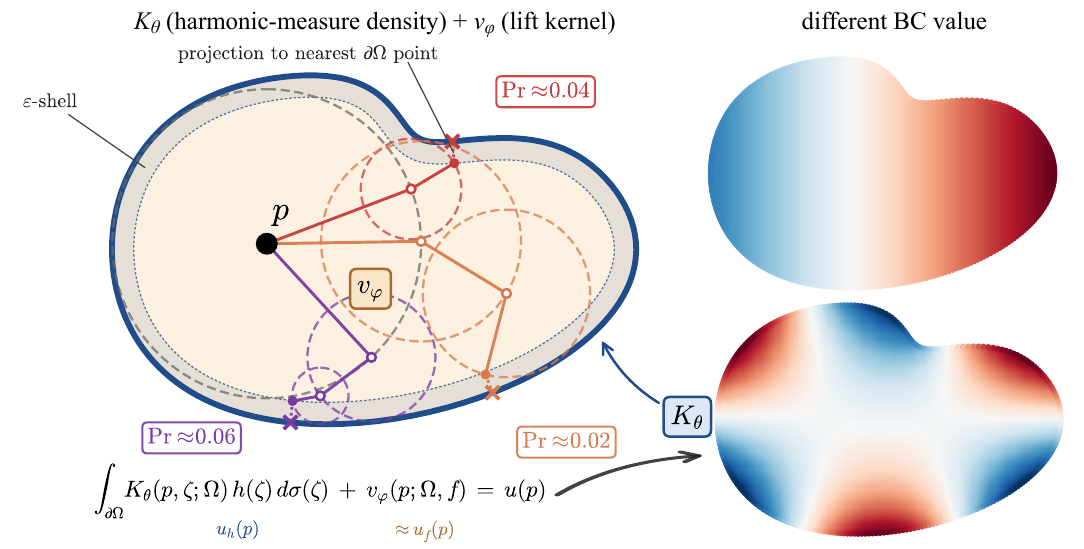}
\caption{\textbf{NHMO overview.} \rev{Harmonic-measure density} $K_\theta \approx \rev{d\omega_p/d\sigma}$ \rev{with WoS paths} and residual lift $v_\varphi$, composed additively \rev{as in~\eqref{eq:nhmo-decomp}} (bottom). \rev{Each WoS step lands on the largest circle inside $\Omega$; a walk stops in the $\varepsilon$-shell (drawn wider than in practice) and is projected to $\partial\Omega$.} Right: $K_\theta$ once fit for $\Omega$ solves any new boundary datum without retraining.}
\label{fig:nhmo-schematic}
\vspace{-1em}
\end{wrapfigure}
$K_\theta(p, \zeta; \Omega)$ is realized as the composition of a geometry encoder $E$ and a kernel head $g$. The encoder maps a discretization of $\Omega$ to a fixed-size shape latent $\psi_\Omega = E(\Omega)$. The kernel head outputs a scalar log-density $\log \rev{\widetilde K_\theta}(p, \zeta; \Omega) = g(p, \zeta, \psi_\Omega)$, with inputs augmented by Fourier features of $(p, \zeta)$ and the inter-point distance $\|p - \zeta\|$. \rev{Given} a boundary discretization $\{\zeta_i\}_{i=1}^{N_s}$ of $N_s$ surface points with quadrature weights $w_i$, \rev{we normalize the kernel over the quadrature at inference, $K_\theta(p,\zeta_i;\Omega) = \widetilde K_\theta(p,\zeta_i;\Omega) / \sum_j w_j \widetilde K_\theta(p,\zeta_j;\Omega)$, so that $\sum_i w_i K_\theta(p,\zeta_i;\Omega) = 1$ holds exactly and} the boundary integral
\begin{equation}
    u_h(p) \;=\; \sum_{i=1}^{N_s} w_i\, K_\theta(p, \zeta_i; \Omega)\, h(\zeta_i)
    \label{eq:bdy-int}
\end{equation}
\rev{is a convex combination of boundary values, so $\min h \le u_h \le \max h$.} A training-time penalty (\S\ref{sec:method-training}) keeps $\sum_i w_i \rev{\widetilde K_\theta}(p, \zeta_i; \Omega) \approx 1$. Encoder, kernel head, and feature parameterizations for the 2D and 3D realizations are in Appendix~\ref{app:architecture}.

By~\eqref{eq:poisson-rep}, when $K_\theta = \rev{d\omega_p/d\sigma}$ and $h$ is analytically harmonic, the boundary integral~\eqref{eq:bdy-int} returns $h(p)$ \rev{up to quadrature error}. We use this to check the trained kernel on held-out geometries by evaluating $u_h$ against analytic harmonic functions (e.g., $h \in \{x,\, xy,\, x^2 - y^2,\, e^x\cos y\}$ in 2D, with low-degree solid spherical harmonics in 3D). The check is unavailable to end-to-end neural operators that do not expose a kernel; numbers are reported in \S\ref{sec:exp-kernel-quality}.

A single fitted $K_\theta$ amortizes solutions across $(h, f)$ pairs on the same shape: for Laplace ($f \equiv 0$) the boundary integral~\eqref{eq:bdy-int} alone suffices; for Poisson the kernel composes additively with $v_\varphi$ for the source contribution. At inference, the discrete effective-kernel matrix $K_{\text{eff}} = [w_j\, K_\theta(p_i,\zeta_j;\Omega)]_{ij}$ is materialized once per geometry and reused for every $(h, f)$, so per-problem inference reduces to a boundary matvec plus a lift forward; wall-clock measurements are in \S\ref{sec:perf}.

\subsection{Field lift $v_\varphi$}\label{sec:method-lift}

\rev{In 2D,} $v_\varphi$ is parameterized as a U-Net on the ambient discretization grid of $\Omega$. Inputs are the interior mask $\mathbf{1}_\Omega$ (the indicator function of $\Omega$, equal to $1$ inside and $0$ outside), the boundary data $h$ extended onto the grid, the source field $f$, and the kernel's own boundary-integral prediction $u_h$ evaluated at every grid pixel. The output is a single residual channel; the final prediction~\eqref{eq:nhmo-decomp} is masked to the interior via multiplication by $\mathbf{1}_\Omega$. The lift sees the kernel's prediction as a guide and learns the residual. For pure Laplace problems ($f \equiv 0$), the kernel alone supplies $u_h$ via~\eqref{eq:bdy-int} and $v_\varphi$ has only the residual approximation error in $K_\theta$ to correct; for Poisson problems, the lift carries the source-induced contribution that the boundary integral cannot represent. Because the lift conditions on $u_h$, the kernel and the lift compose into a single forward pass per query field with no iterative coupling. \rev{In 3D, $v_\varphi$ is a cross-attention head whose query is a Fourier embedding of $p$ and whose context is the shape latent together with tokens that summarize source samples $(q_j, f(q_j))$; it does not see $h$, and its output is multiplied by $\max(0, -\mathrm{SDF}(p))$.} The same kernel-plus-lift template adapts to nearby elliptic operators, e.g.\ constant-drift Laplace, by replacing the lift with a small drift-conditioned adapter while reusing the geometric kernel without retraining; we demonstrate this on a bunny domain in \S\ref{sec:exp-intuitive}. \rev{Depths, widths, and parameter counts} are in Appendix~\ref{app:architecture}.

\subsection{Training}\label{sec:method-training}

Training is two-stage. The kernel $K_\theta$ is trained per geometry distribution from Walk-on-Spheres exit samples~\citep{muller1956some,sawhney2020monte}. From each interior probe $p$, we simulate $M$ Brownian-motion exit points $\{\zeta_k\}_k \subset \partial\Omega$\rev{: each WoS step jumps to a uniform point on the largest sphere around the current point inside $\Omega$, a walk stops in the $\varepsilon$-shell of $\partial\Omega$ ($\varepsilon = 10^{-3}$ of the normalized domain) and is projected to the nearest boundary point, and walks that do not stop within $128$ steps are masked out. In 2D, we precompute $10^4$ walks for each of $32$ probes per shape, and in 3D we draw $4$ fresh exits for each of $8$ probes per gradient step. We} fit $K_\theta(p,\cdot)$ against a Gaussian kernel-density estimate \rev{(KDE)} of these samples at bandwidth $\sigma$ (a small fraction of the domain diameter\rev{, $0.2\%$ in 2D, so $\varepsilon$ is half of $\sigma$}), minimizing the KDE negative log-likelihood. \rev{Probes are drawn from near-boundary, mid-interior, and deep-interior bands.} Two regularizers harden the soft normalization. $\mathcal{L}_Z$ pins $\log \sum_i w_i \rev{\widetilde K_\theta}(p,\zeta_i;\Omega)$ to zero via a Huber penalty, and $\mathcal{L}_{\text{MV}}$ enforces the mean-value property of harmonic functions on spheres $B(p, r) \subset \Omega$ that lie strictly inside $\Omega$. \rev{Generating this supervision is a negligible share of training: our GPU sampler completes about $5\times 10^8$ walks per second on an A100 even at a stricter $\varepsilon = 10^{-4}$ (Appendix~\ref{app:wos-budget} gives budgets, masked fractions, and variance).}

With $K_\theta$ frozen, the lift $v_\varphi$ is trained by masked MSE between the composed prediction~\eqref{eq:nhmo-decomp} and the numerical reference $u_{\text{true}}$, computed in $y$-normalized space. One shape $\times$ one $(h, f)$ instance per gradient step; $u_h$ is recomputed on-the-fly through the frozen kernel. No PDE-residual loss is used at any stage. Loss weights, optimizer, and learning-rate schedule for both stages are in Appendix~\ref{app:losses}.

\section{Experiments}\label{sec:experiments}
\begin{figure}[t]
    \centering
    \includegraphics[width=\linewidth]{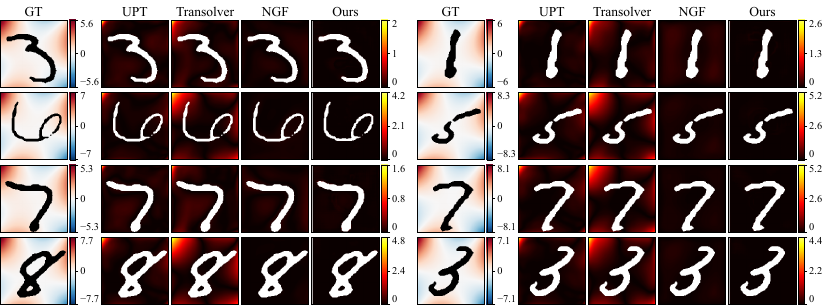}
    \caption{\textbf{2D MNIST out-of-distribution (OOD) qualitative.} Per row, a Laplace example (left) paired with a Poisson example (right); columns are GT \rev{(finite-difference reference)} and \rev{the} absolute error \rev{$|\text{pred} - \text{GT}|$} of \rev{UPT, Transolver, NGF, and} Ours. \rev{Within each example (half-row), the four error panels share one color scale and GT has its own.} Additional shapes in Appendix~\ref{app:mnist-extra}.}
    \label{fig:mnist_qualitative}
\end{figure}
We evaluate NHMO on two complementary benchmarks, a controlled 2D MNIST testbed in which all neural-operator baselines run under a single code path (\S\ref{sec:exp-mnist}) and the published 3D MCB-B Poisson benchmark of~\citep{yoo2025neural} where NHMO is compared against four prior methods on five mechanical-part categories (\S\ref{sec:exp-mcb}), preceded by a short pair of intuitive demonstrations on complex 3D shapes (\S\ref{sec:exp-intuitive}). \rev{Runtime (\S\ref{sec:perf}) and ablation (\S\ref{sec:ablation}) analyses follow.}

\subsection{Intuitive demonstrations on complex shapes}\label{sec:exp-intuitive}

As proof of concept, two demos share a single harmonic-measure \rev{density} $K_\theta$ fitted once across four graphics meshes; under matched optimization budgets, NHMO converges faster than a Green's-function-style baseline (GF style) that learns a volumetric Green's function as in prior work~\citep{yoo2025neural,boulle2022data,teng2022learning,negi2024learning,gin2021deepgreen,li2020multipole,teixeira2026variational}. \rev{\emph{(i)} On four graphics meshes (armadillo, bunny, fandisk, lucy) with $h \in \{\sin x,\,\sin z\}$, NHMO reaches mean rel-$L_2$ of $0.012$ against an FEM reference versus $0.142$ for GF style. \emph{(ii)} For constant-drift Laplace $\Delta u + \beta\cdot\nabla u = 0$ on a 2D bunny slice, the same $K_\theta$ plus a small drift-conditioned adapter reaches $0.062$ versus $0.323$ for GF style. Details and figures are in Appendix~\ref{app:intuitive}.}

\subsection{2D MNIST: controlled cross-baseline benchmark}\label{sec:exp-mnist}

We construct a controlled 2D testbed using MNIST digit silhouettes as planar domains, with parametric Laplace and Poisson problems posed on each, and run all neural-operator baselines under one training and evaluation pipeline. \rev{It probes out-of-distribution (OOD) extrapolation across BC coefficients and multiply-connected boundaries (digits 0, 6, 8, 9); details are in Appendix~\ref{app:mnist-setup}.} \rev{The baselines include BENO~\citep{wang2024beno}, which is designed for elliptic problems with complex boundaries.}

\begin{revblock}
Table~\ref{tab:mnist_main} reports results on the in-distribution and OOD splits; the OOD split draws BC coefficients strictly outside the training range. NHMO has the lowest mean in-distribution error and the lightest error tails on both splits (OOD examples in Figure~\ref{fig:mnist_qualitative}). Under the OOD shift it degrades by $1.25{\times}$ (median), while the nonlinear end-to-end baselines (Transolver, LNO, UPT, BENO) degrade by $4{\times}$ to $8{\times}$; even the kernel-only variant beats all of them on OOD. Our 2D port of NGF also extrapolates well and has a quite low OOD mean and median: like NHMO, it pairs geometry-only features with a read-out that is linear in the data, the class of operator this paper argues for. Its in-distribution errors, however, are heavy-tailed on Poisson problems (p95 $16.9\%$ and max $40.2\%$, against our $3.3\%$ and $7.0\%$), consistent with its rank-limited bilinear source coupling. Across five training seeds of the lift, the test mean is $2.09 \pm 0.03\%$ and the OOD mean $2.60 \pm 0.05\%$ (Appendix~\ref{app:ladder}).
\end{revblock}

\begin{table}[h]
\centering
\small
\setlength{\tabcolsep}{3.5pt}
\caption{2D MNIST paramBC: relative $L_2$ \rev{error (\%)} over the full interior, in-distribution and OOD ($U[+1,+2]$). \rev{p95: per-pair $95$th percentile.} The OOD/test ratio \rev{(of medians)} measures BC-coefficient extrapolation; lower is better. Per-problem inference is on a single A100 with the per-shape kernel matrix cached. \rev{NGF is released only in 3D; its row is our 2D port of the official code and training setup (Appendix~\ref{app:ngf-2d}).}}
\label{tab:mnist_main}
\begin{tabular}{lccccccc}
\toprule
 & \multicolumn{2}{c}{test (in-dist)} & \multicolumn{2}{c}{test\_ood} & & \\
\cmidrule(lr){2-3}\cmidrule(lr){4-5}
Method & \rev{median / mean} & \rev{p95} & \rev{median / mean} & \rev{p95} & OOD/test & per-problem \\
\midrule
Transolver~\citep{wu2024transolver}  & $4.4$ / $5.3$ & \rev{$10.7$} & $36.4$ / $36.2$ & \rev{$55.2$} & $8.3{\times}$ & $12.5$\,ms \\
LNO~\rev{\citep{wang2024latent}}           & $5.2$ / $6.4$ & \rev{---} & $23.3$ / $33.2$ & \rev{$81.6$} & $4.5{\times}$ & $13.2$\,ms \\
UPT~\citep{alkin2024universal}       & $5.7$ / $6.3$ & \rev{---} & $26.5$ / $26.6$ & \rev{---} & $4.6{\times}$ & $7.6$\,ms \\
\rev{BENO~\citep{wang2024beno}} & \rev{$5.9$ / $7.0$} & \rev{$16.1$} & \rev{$44.1$ / $40.5$} & \rev{$65.8$} & \rev{$7.5{\times}$} & \rev{---} \\
NGF~\citep{yoo2025neural} \rev{(2D port)} & \rev{$2.0$ / $3.9$} & \rev{$16.9$} & \rev{${3.9}$ / ${4.2}$} & \rev{$6.0$} & \rev{${1.93{\times}}$} & $11.6$\,ms \\
\midrule
\textbf{NHMO K-only} (kernel only)   & $5.4$ / $7.7$ & \rev{$18.2$} & $5.6$ / $6.8$ & \rev{$14.2$} & $1.0{\times}$ & --- \\
\textbf{NHMO} (kernel + lift) & $2.0$ / $2.1$ & \rev{$3.3$} & $2.5$ / $2.6$ & \rev{$4.2$} & $1.25{\times}$ & $\mathbf{4.0}$\,\textbf{ms} \\
\rev{\quad + residual head (\S\ref{sec:discussion})} & \rev{$\mathbf{1.7}$ / $\mathbf{1.8}$} & \rev{$\mathbf{3.0}$} & \rev{$\mathbf{2.4}$ / $\mathbf{2.6}$} & \rev{$\mathbf{4.1}$} & \rev{$1.37{\times}$} & \rev{---} \\
\bottomrule
\end{tabular}
\end{table}

\subsection{3D MCB-B Poisson}\label{sec:exp-mcb}
\begin{table}
\centering
\caption{MCB-B Poisson benchmark: relative $L_2$ error against the FEM reference, mean over $20$ unseen test shapes $\times$ $16$ unseen $(h,f)$ problems per category. Lower is better. NGF, Transolver, LNO, UPT numbers are reported in NGF Table~2 under identical evaluation.}
\label{tab:mcb_main}
\begin{tabular}{lccccc}
\toprule
Method & Nut & Gear & Motor & Fitting & Screws \& Bolts \\
\midrule
Transolver~\citep{wu2024transolver}      & 0.320 & 0.281 & 0.407 & 0.180 & 0.221 \\
LNO~\rev{\citep{wang2024latent}}               & 0.372 & 0.466 & 0.528 & 0.259 & 0.239 \\
UPT~\citep{alkin2024universal}           & 0.516 & 0.507 & 0.765 & 0.392 & 0.358 \\
NGF~\citep{yoo2025neural}                & 0.275 & 0.243 & 0.338 & 0.160 & 0.189 \\
\midrule
\textbf{Ours (NHMO)}                     & \textbf{0.216} & \textbf{0.188} & \textbf{0.284} & \textbf{0.147} & \textbf{0.131} \\
\bottomrule
\end{tabular}
\end{table}
We follow the protocol of~\citep{yoo2025neural} exactly. MCB-B comprises five categories of mechanical-part shapes (Nut, Gear, Motor, Fitting, Screws \& Bolts) from MCB~\citep{kim2020large}, each with 200 training and 20 unseen test shapes; per shape, the benchmark provides 16 unseen $(h,f)$ problems (8 sources $\times$ 2 BCs, held out from training) with FEM reference solutions to $\Delta u = f$ on tetrahedral meshes, yielding 320 test pairs per category. Our networks ($K_\theta$ at $2.77$M params trained with WoS distillation, $v_\varphi$ at $\sim 5$M params trained on FEM-supervised MSE via warm-start, \S\ref{sec:method-training}) and baseline implementations are detailed in Appendices~\ref{app:architecture} and~\ref{app:baselines}. Reported metric is relative $L_2$ error against the FEM reference, evaluated at every interior \rev{tetrahedral-mesh (tet)} vertex and averaged over all $320$ test pairs.

We outperform NGF on all five categories and outperform Transolver, LNO, and UPT by wider margins (Table~\ref{tab:mcb_main}). Per-shape distributions (Table~\ref{tab:mcb_distrib}) show median error below NGF's reported mean for all five categories, with $95$th-percentile error below $0.50$ on every category, so no single test shape fails catastrophically. \rev{When $h$ and $f$ are shifted outside their training ranges without retraining ($40$ problems per category; Appendix~\ref{app:ood3d}), the macro-averaged error of the released NGF checkpoints rises from $0.241$ (Table~\ref{tab:mcb_main}) to $0.615$, and ours from $0.193$ to $0.263$ on the same problems. On a Laplace-only track with the same boundary data, ours averages $0.099$ against $0.60$--$0.64$ for NGF, which suggests that most of our remaining degradation comes from the source-conditioned lift.}

\begin{figure}[t]
    \centering
    \includegraphics[width=0.95\linewidth]{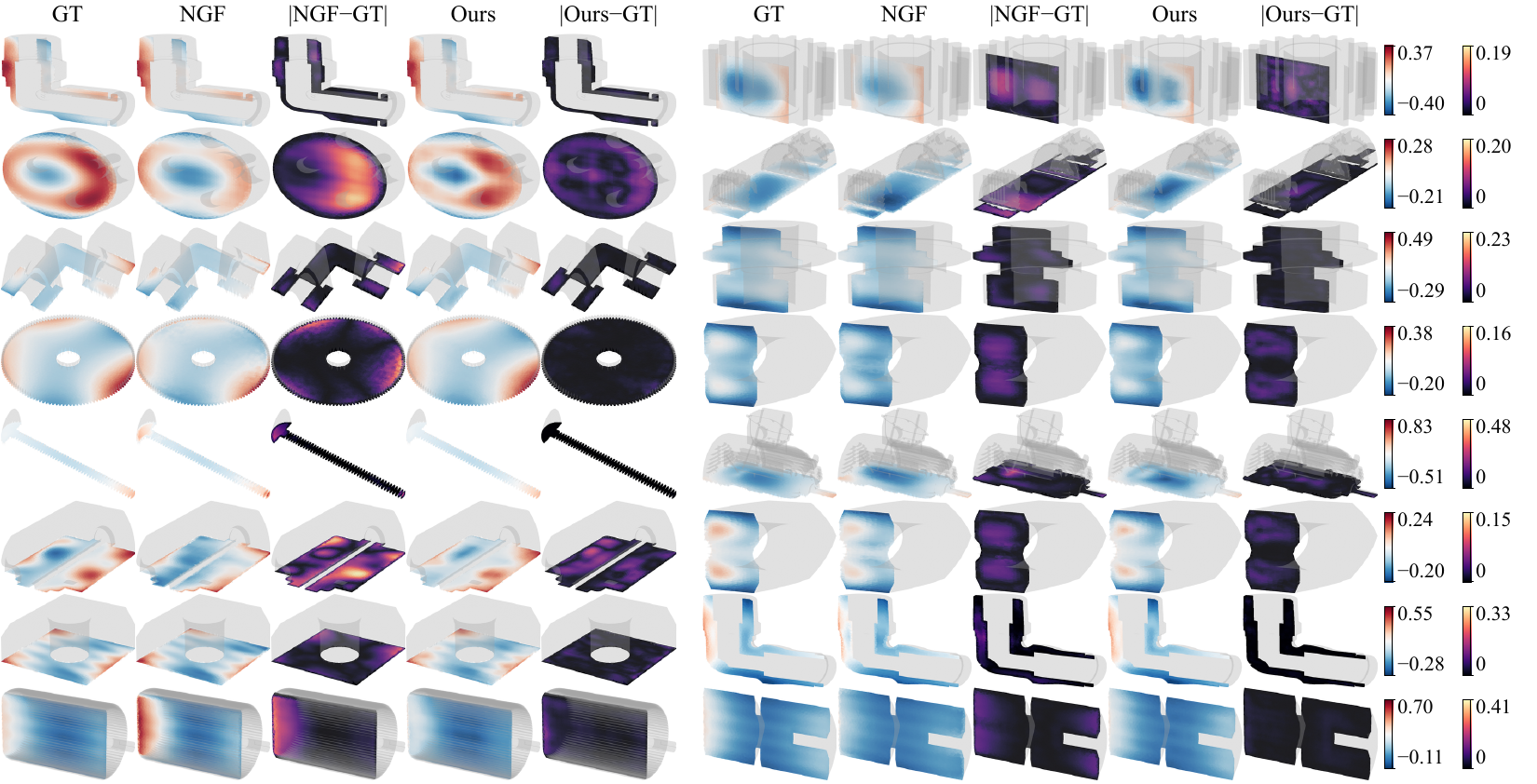}
    \caption{\textbf{Qualitative comparison on MCB-B Poisson.} Per shape, five panels show GT \rev{(the FEM reference)}, NGF prediction, NGF error, our prediction, and our error; the cut face is colored by the field, the back half by a gray ghost surface. \rev{Within each row, GT and the predictions share one color scale and the four error panels share a second one; panels are stretched to a common aspect ratio.} Two shapes per row across all five MCB-B categories. Additional shapes in Appendix~\ref{app:fig4-extra}.}
    \label{fig:qualitative}
\end{figure}

\begin{table}[t]
\centering
\footnotesize
\setlength{\tabcolsep}{3pt}
\begin{minipage}[t]{0.44\linewidth}
\centering
\caption{Per-shape distribution of relative $L_2$ error (ours, $320$-pair test set).}
\label{tab:mcb_distrib}
\begin{tabular}{lccc}
\toprule
 & Mean & Median & p95 \\
\midrule
Nut     & 0.216 & 0.215 & 0.329 \\
Gear    & 0.188 & 0.144 & 0.378 \\
Motor   & 0.284 & 0.265 & 0.450 \\
Fitting & 0.147 & 0.111 & 0.309 \\
Screws  & 0.131 & 0.103 & 0.315 \\
\bottomrule
\end{tabular}
\end{minipage}\hfill
\begin{minipage}[t]{0.53\linewidth}
\centering
\caption{Synthetic Laplace probe: relative $L_2$ error of $u_h(p)=\langle h, K_\theta(p,\cdot)\rangle$ against analytical $u_h(p)\equiv h(p)$, mean over 20 unseen test shapes per category, 64 interior queries per shape.}
\label{tab:kernel_quality}
\begin{tabular}{lcccccc}
\toprule
 & $x$ & $xy$ & $x^2{-}y^2$ & $Y_{2,0}$ & $e^x\!\cos y$ & $e^x\!\sin y$ \\
\midrule
Nut     & 0.149 & 0.190 & 0.172 & 0.171 & 0.043 & 0.094 \\
Gear    & 0.048 & 0.089 & 0.090 & 0.097 & 0.021 & 0.068 \\
Motor   & 0.121 & 0.184 & 0.216 & 0.213 & 0.044 & 0.149 \\
Fitting & 0.091 & 0.167 & 0.161 & 0.153 & 0.029 & 0.118 \\
Screws  & 0.064 & 0.178 & 0.112 & 0.110 & 0.025 & 0.218 \\
\bottomrule
\end{tabular}
\end{minipage}
\end{table}
\paragraph{Kernel as a harmonic-measure density.}\label{sec:exp-kernel-quality} A direct test of whether $K_\theta$ approximates the true harmonic-measure \rev{density} is to evaluate its boundary integral against \emph{analytically harmonic} $h$, where $u_h(p) \equiv h(p)$ exactly by uniqueness of the harmonic extension. Table~\ref{tab:kernel_quality} reports rel-$L_2$ errors for $h\in\{x,\,xy,\,x^2-y^2,\,Y_{2,0},\,e^x\cos y,\,e^x\sin y\}$ ($Y_{2,0}$ is the degree-2 zonal solid spherical harmonic) across all five categories. Errors are small and ordered consistently with a true harmonic-measure density (smoothest probes lowest, second-order spherical harmonics highest), and the ordering is preserved on the harder Motor and Fitting geometries\rev{; part of the remaining Poisson error (Table~\ref{tab:mcb_main}) therefore stems from the source lift}.

\begin{revblock}
\subsection{Runtime}\label{sec:perf}

Every method first turns a new geometry into the representation it computes on. For NHMO, this geometry step encodes the shape and evaluates $K_{\text{eff}}$ once, playing the role that meshing plays for mesh-based pipelines (Table~\ref{tab:runtime}).

\begin{table}[h]
\centering
\small
\setlength{\tabcolsep}{4pt}
\caption{Runtime on a single A100. The geometry step runs once per shape: shape encoding and $K_{\text{eff}}$ for NHMO, tetrahedral meshing at the released resolution (fTetWild, CPU) for the MCB-B baselines.}
\label{tab:runtime}
\begin{tabular}{lcccc}
\toprule
 & \multicolumn{2}{c}{Geometry step (per shape)} & \multicolumn{2}{c}{Per problem} \\
\cmidrule(lr){2-3}\cmidrule(lr){4-5}
Setting & NHMO & baselines & NHMO & baselines \\
\midrule
2D MNIST ($128^2$) & $6.6$\,s & grid input & $4.0$\,ms & $7.6$--$13.2$\,ms \\
3D Nut   & $7.9$\,s & $37$\,s (tet meshing) & $7.5$\,ms & $0.24$\,s (NGF) \\
3D Motor & $10.3$\,s & $48$\,s (tet meshing) & $7.9$\,ms & $0.25$\,s / $77$\,ms (NGF) \\
\bottomrule
\end{tabular}
\end{table}

FEM and all MCB-B baselines, including NGF, take as input the vertices of the tetrahedral mesh that the benchmark provides. On a new shape, fTetWild~\citep{hu2020fast} takes $37$/$48$\,s to mesh the Nut/Motor surfaces at the released resolution, while our geometry step takes $7.9$/$10.3$\,s from the boundary surface alone, so NHMO is faster than the mesh-based pipelines from the first problem. NGF's formulation does not use mesh connectivity, but any other interior point set would also require a comparable geometry step. Per problem, NHMO solves in $7.5$/$7.9$\,ms, against $0.24$/$0.25$\,s for NGF's released pipeline (including data loading) and $77$\,ms per forward pass when the Motor mesh is preloaded on the GPU and only the boundary data change. In 2D, our geometry step takes $6.6$\,s per shape. Workloads that query one geometry many times, such as parametric boundary-condition studies, load sweeps on a fixed part, and uncertainty quantification, benefit the most: sweeping $1{,}000$ load cases on one Motor shape takes NHMO about $18$\,s, geometry step included, against about $77$\,s for NGF with a preloaded mesh. Because the geometry step depends only on the shape, it can also be run ahead of time for a library of shapes. The 3D timings use inference-only optimizations whose effect on accuracy is within sampling noise (Appendix~\ref{app:speed}).
\end{revblock}

\subsection{Ablations}\label{sec:ablation}
The full table and per-ablation discussion are in Appendix~\ref{app:ablations}.

\textbf{(1) Accuracy is independent of geometric representation.} Swapping the geometry encoder between a point-cloud over boundary samples and a 2D SDF image moves median rel-$L_2$ by only $0.27\%$ on test and $0.09\%$ on OOD, and the OOD/test gap tightens to $1.14\times$ (from canonical $1.25\times$). Setup in Appendix~\ref{app:abl-encoder}.
\textbf{(2) Mixed-corpus generalization across all 10 MNIST classes.} A single $K_\theta + v_\varphi$ trained on a $5{,}000$-shape corpus across digits $0$--$9$ attains a per-class mean spread of only $0.53\%$ \rev{(max $-$ min, test)}.
\textbf{(3) Factorization, not the lift.} The kernel-only variant ($v_\varphi \equiv 0$) already beats \rev{the nonlinear end-to-end baselines} on OOD (Table~\ref{tab:mnist_main}); the lift is a small correction.
\textbf{(4) \rev{Robustness to the boundary quadrature}.} Varying \rev{$n_{\text{surf}}$ from $50$ to $400$ changes the median by ${<}0.1\%$ above $100$ samples.}
\textbf{(5) Not tuned on a knife-edge.} Doubling lift parameters from $6.4$M to $11.3$M gives no in-distribution gain; KDE $\sigma$ is robust across a $5\times$ range; WoS supervision converges above ${\sim}1{,}000$ samples per query.
\begin{revblock}
\section{Discussion}\label{sec:discussion}

\paragraph{Why a boundary density and an amortized source field.}\label{sec:why-not-green} Green's-function operators such as NGF learn one kernel $G_\Omega(p,q)$ and integrate it against $f$ in the volume and its normal derivative against $h$ on the boundary. We learn the boundary density and the integrated source field instead, for three reasons. \emph{Supervision}: WoS exit points sample $\omega_p$, so $K_\theta$ is trained from walks alone, whereas learned-$G$ methods rely on solver-generated solution fields. \emph{Boundary accuracy}: near $\partial\Omega$ the value of $G_\Omega$ vanishes and the signal sits in its normal derivative, so a learned $G$ must be accurate enough to be differentiated there. \emph{Inference cost}: a learned $G$ needs a new singular volume quadrature, $O(N_p N_q)$ network evaluations, whenever $f$ changes. In a direct test (Appendix~\ref{app:gvol}), a learned volumetric integrand with a $\log|p-q|$ singularity feature reaches $6.7\%$ / $5.2\%$ (mean / median), no better than a source-only field lift, and its boundary derivative is not a valid Poisson kernel. NGF makes this integral fast with a rank-constrained bilinear factorization, and its heavy error tails (\S\ref{sec:exp-mnist}) are consistent with that rank limit.

\paragraph{The boundary-data dependence of the lift.} In the classical balayage split the source-only piece does not depend on $h$, whereas our 2D lift sees $h$ and $u_h$, because $K_\theta$ is a fitted density: the boundary term leaves the residual $e_h(p) = \int_{\partial\Omega} h\,(d\omega_p/d\sigma - K_\theta)\,d\sigma$, a linear functional of $h$ that a network seeing only $(\Omega, f)$ cannot correct. On the $205$ Laplace test pairs the source contribution vanishes and the output of the lift is its boundary correction alone: it correlates with $e_h$ at median $0.97$ and removes $71\%$ of it (Appendix~\ref{app:ladder}). The two roles can also be separated into a source lift $v_\varphi(\Omega, f)$, which sees only the mask, its SDF, and $f$, and a residual head $r(\Omega, h, u_h)$ trained on top of it, giving $u = \langle h, K_\theta\rangle + v_\varphi(\Omega, f) + r(\Omega, h, u_h)$. The source lift alone reaches $6.1\%$ / $4.5\%$ (test mean / median) and degrades only $1.04{\times}$ under the OOD shift; adding $r$ gives $1.84\%$ / $1.74\%$ on test and $2.56\%$ / $2.39\%$ on OOD, which matches or slightly surpasses the two-term model ($2.1\%$ / $2.0\%$ and $2.6\%$ / $2.5\%$) at a larger total capacity, with the $h$-dependence confined to an explicit corrector. The 3D lift sees only the geometry and the source, as in the classical split. The residual comes mainly from the training signal of the 2D kernel, whose KDE targets are built on simplified polyline contours rather than on the rasterized masks (Appendix~\ref{app:residual-origin}); placing the KDE nodes on the mask contour lowers the kernel-only error from $5.9\%$ to $3.2\%$, and a kernel-plus-lift model trained on this kernel reaches $1.8\%$ / $1.7\%$ on test and $2.1\%$ / $2.2\%$ on OOD; we leave this choice of representation, and residual heads that exploit the linearity of $e_h$, to future work.
\end{revblock}

\section{Conclusion}\label{sec:conclusion}

We proposed Neural Harmonic Measure Operator (NHMO), the first neural operator built explicitly around the harmonic measure, the canonical probability distribution from potential theory that mediates all solutions of the Dirichlet Laplace problem on a fixed domain. For Poisson source terms, the classical balayage decomposition extends the same harmonic measure to handle sources via a learned lift network with a zero-boundary gauge. Our framework reduces an end-to-end neural-operator problem to two coupled components: the boundary kernel $K_\theta$, independently falsifiable as a harmonic-measure \rev{density} via synthetic-harmonic probes, and the amortized balayage source lift $v_\varphi$. More broadly, our work suggests that grounding neural operators in canonical objects from classical analysis, rather than learning end-to-end input-to-output mappings, offers a path to inductive biases that mirror the structure of the underlying PDE, and we hope this framing motivates further work at the interface of potential theory and neural operator learning.
\paragraph{Limitations and future work.} \rev{We target Dirichlet elliptic problems; Neumann/Robin conditions and other PDE types need generalized measures and remain future work. The source lift is $f$-conditional via probe samples, so out-of-distribution sources may degrade. Linearity in $h$ is guaranteed only for the kernel channel: a lift that learns shortcuts specific to the training coefficients would lose OOD robustness.} NHMO also trades a per-shape precompute for fast per-problem inference, so single-problem-per-shape workloads do not benefit from the cache; future work could explore low-rank kernel factorizations to amortize this cost.

\begin{ack}
We sincerely thank the reviewers for their valuable feedback. Georgia Tech authors acknowledge NSF CAREER \#2420319, IIS \#2433307, OISE \#2433313, IIS \#2433322, ECCS \#2318814, and CNS \#2450401 for funding support. We thank NVIDIA for providing computing resources through the NVIDIA Academic Grant. The authors declare no competing interests.
\end{ack}

\bibliography{references,references_cr}
\bibliographystyle{plainnat}

\clearpage
\appendix

\startcontents[appendix]
\section*{Appendix}

\printcontents[appendix]{}{0}{\setcounter{tocdepth}{2}}
\clearpage
\section{Notation}\label{app:notation}

\begin{table}[h]
\centering
\small
\caption{Notation used throughout the paper.}
\label{tab:notation}
\begin{tabular}{ll}
\toprule
Symbol & Meaning \\
\midrule
\multicolumn{2}{l}{\emph{Geometry}} \\
$\Omega \subset \R^d$ & bounded Lipschitz domain in dimension $d \in \{2,3\}$ \\
$\partial\Omega$ & boundary of $\Omega$ \\
$p, q \in \Omega$ & interior points \\
$\zeta \in \partial\Omega$ & boundary point \\
$\nu_\zeta$ & outward unit normal to $\partial\Omega$ at $\zeta$ \\
$d\sigma$ & surface measure on $\partial\Omega$ \\
$\mathrm{SDF}(p)$ & signed distance to $\partial\Omega$ (negative inside $\Omega$) \\
\midrule
\multicolumn{2}{l}{\emph{PDE data}} \\
$h \in C(\partial\Omega)$ & Dirichlet boundary data \\
$f \in L^\infty(\Omega)$ & source term in the Poisson problem $\Delta u = f$ \\
$u$ & PDE solution \\
\midrule
\multicolumn{2}{l}{\emph{Classical potential-theoretic objects}} \\
$\omega_p$ & harmonic measure at $p$ (probability measure on $\partial\Omega$) \\
\rev{$d\omega_p/d\sigma$} & \rev{harmonic-measure density (Poisson kernel), equal to $-\partial_\nu G_\Omega(p,\cdot)$} \\
$G_\Omega(p,q)$ & Dirichlet Green's function on $\Omega$ \\
$\Phi$ & fundamental solution of $-\Delta$ on $\R^d$ \\
$N_f$ & Newtonian potential of $f$ \\
$u_f$ & particular Poisson solution with $u|_{\partial\Omega}=0$ \\
$B_t$, $\tau$ & Brownian motion in $\R^d$, first-exit time from $\Omega$ \\
\midrule
\multicolumn{2}{l}{\emph{Learned objects}} \\
$K_\theta(p,\zeta;\Omega)$ & learned harmonic-measure density (NHMO kernel\rev{, quadrature-normalized}) \\
$\widetilde K_\theta$ & pre-normalization network output (cf.\ soft normalization) \\
$v_\varphi(p;\Omega,f)$ & learned source lift (zero-boundary-gauge) \\
\rev{$r(p;\Omega,h,u_h)$} & \rev{learned residual head (2D)} \\
\rev{$e_h(p)$} & \rev{kernel-fit residual $\int_{\partial\Omega} h\,(d\omega_p/d\sigma - K_\theta)\,d\sigma$} \\
$\mathrm{sl}(\Omega)$ & learned shape latent \\
$u_h(p)$ & boundary integral $\sum_i w_i\,K_\theta(p,\zeta_i;\Omega)\,h(\zeta_i)$ \\
\midrule
\multicolumn{2}{l}{\emph{Discretization}} \\
$\{\zeta_i\}_{i=1}^{N_s}$ & surface quadrature samples on $\partial\Omega$ \\
$\{w_i\}_{i=1}^{N_s}$ & surface quadrature weights ($w_i \propto \sigma(\partial\Omega)/N_s$) \\
$\{q_j\}_{j=1}^{N_p}$ & interior source-probe samples \\
\midrule
\multicolumn{2}{l}{\emph{Abbreviations}} \\
OOD & out-of-distribution (evaluation split with BC coefficients outside training) \\
\rev{KDE} & \rev{kernel-density estimate (of WoS exit points)} \\
\rev{GT / GF} & \rev{ground truth (numerical reference) / Green's-function-style baseline} \\
\bottomrule
\end{tabular}
\end{table}

\section{Harmonic measure: derivations and properties}\label{app:harmonic-measure}

This appendix collects the classical facts about $\omega_p$ deferred from~\S\ref{sec:bg-harmonic}.

\paragraph{Existence/uniqueness and Riesz representation.} For $h \in C(\partial\Omega)$, the Dirichlet problem
\begin{equation}\label{eq:dirichlet}
    \Delta u = 0 \text{ in } \Omega, \qquad u = h \text{ on } \partial\Omega
\end{equation}
admits a unique solution $u \in C(\overline{\Omega}) \cap C^2(\Omega)$ on a bounded Lipschitz domain. Linearity in $h$ together with the maximum principle make $h \mapsto u(p)$ a positive linear functional on $C(\partial\Omega)$ for each fixed $p \in \Omega$. The Riesz representation theorem then yields a unique probability measure $\omega_p$ on $\partial\Omega$ such that $u(p) = \int_{\partial\Omega} h\, d\omega_p$, recovering~\eqref{eq:poisson-rep}~\citep{garnett2005harmonic}. Positivity and total mass one are automatic from this construction, and combined with~\eqref{eq:poisson-rep} they give the maximum principle $\min h \le u \le \max h$.

\paragraph{Green's-function trace formula.} When $\partial\Omega$ is sufficiently regular (smooth, $C^1$, or more generally Lipschitz~\citep{garnett2005harmonic}), $\omega_p$ is absolutely continuous with respect to surface measure $d\sigma$, and its Radon--Nikodym density coincides $\sigma$-almost everywhere with the (negated) nontangential outward-normal derivative of the Dirichlet Green's function:
\begin{equation}\label{eq:green-trace}
    \frac{d\omega_p}{d\sigma}(\zeta) \;=\; -\partial_{\nu_\zeta} G_\Omega(p, \zeta) \quad \text{($\sigma$-a.e.\ on $\partial\Omega$)},
\end{equation}
where $\nu_\zeta$ is the outward unit normal at $\zeta$. The right-hand side is non-negative because $G_\Omega$ is positive in $\Omega$ and zero on $\partial\Omega$. \rev{The kernel $K_\theta$ approximates this density, so $K_\theta\,d\sigma$ approximates $\omega_p$, and the quadrature weights $w_i$ in~\eqref{eq:bdy-int} discretize $d\sigma$.}

\begin{revblock}
\paragraph{Walk-on-Spheres as a sampler of $\omega_p$.} For the isotropic Laplacian, Brownian motion started at the center of a ball contained in $\Omega$ leaves the ball at a uniformly distributed point of its sphere (the mean-value property). WoS chains such jumps, each on the largest sphere around the current point that fits in $\Omega$, so an ideal walk draws its exit point exactly from $\omega_p$, and the average of $h$ over exit points has expectation $u(p) = \mathbb{E}_p[h(B_\tau)]$ for any number of walks~\citep{muller1956some}. WoS samples the exit law directly and integrates no density, so the surface measure enters only when the learned density is integrated with the quadrature weights $w_i$. The implemented walk carries three small biases: the $\varepsilon$-shell termination, whose bias is $O(\varepsilon)$ on our Lipschitz domains~\citep{mascagni2003epsilon}; the step cap, with non-terminating walks masked out (their fraction is reported in Appendix~\ref{app:wos-budget}); and the rasterized SDF used to compute sphere radii. The expected number of steps grows as $O(\log(1/\varepsilon))$ with geometry-dependent constants~\citep{binder2012rate}. The uniform-sphere jump relies on the Euclidean, isotropic Laplacian; drift or varying coefficients need transformed walks, and our drift demonstration (Appendix~\ref{app:intuitive-drift}) uses a Yukawa-type transform.
\end{revblock}

\paragraph{Further properties.} In 2D, $\omega_p$ is invariant under conformal maps of $\Omega$~\citep{garnett2005harmonic}. Its dimensional properties characterize boundary regularity~\citep{makarov1985distortion}. Neither property is used in our construction; we record them only for completeness.

\section{Architecture details}\label{app:architecture}

This section gives concrete dimensions for the canonical 2D MNIST configuration\rev{, followed by the 3D MCB-B configuration}.

\paragraph{Shape encoder.} A Transolver-style slice-attention module. Boundary samples (point + outward normal) are augmented with Fourier features ($L=8$ bands per axis) and a learned boundary / interior-anchor type embedding, then projected to $d_{\text{model}}=256$ tokens. A soft slice projection produces $M=64$ slice tokens (independent of the input boundary density), followed by a $3$-layer pre-norm transformer encoder with $4$ heads and MLP ratio $4$. The SDF variant referenced in \S\ref{app:abl-encoder} replaces the point-cloud stem with a $3$-block CNN over a $64\times 64$ SDF grid that yields a $16\times 16$ token grid (also $d_{\text{model}}=256$); all downstream hyperparameters are held identical.

\paragraph{Kernel head.} Cross-attention with $n=2$ pre-norm layers, $4$ heads, $d_{\text{model}}=256$, MLP ratio $4$. The query token is built from Fourier features of $p$ ($L=10$). Boundary tokens are queries; their inputs are Fourier features of $\zeta$ ($L=10$) plus the outward normal ($L=4$) and the displacement $\Delta = \zeta - p$ ($L=4$ plus the raw vector). The cross-attention context is the encoder output $Z(\Omega)$ concatenated with the query token. Read-out is a $2$-layer MLP onto a scalar logit per boundary sample, normalized via softmax weighted by the surface quadrature weights so that $\sum_i w_i\, K_\theta(p, \zeta_i;\Omega) \rev{= 1}$. Logit clipping (the $\log K_{\max}$ cap of earlier configurations) is disabled in the canonical configuration.

\paragraph{Field lift.} A symmetric 2D U-Net with input channels $(\mathbf{1}_\Omega, h, f, u_h)$, base width $48$, depth $4$, GroupNorm ($8$ groups), GELU. Three down-blocks take channels $48\to 96\to 192\to 384$ at spatial resolutions $128\to 64\to 32\to 16$; a middle block; three up-blocks with skip connections; a $1\times 1$ output projection. The output is multiplied by the interior mask. \rev{The 3D lift is described below. The variant of \S\ref{sec:discussion} replaces the field lift by a source lift with input channels $(\mathbf{1}_\Omega, \mathrm{SDF}, f)$ and base width $64$ (${\approx}11.3$M parameters) and a residual head with input channels $(\mathbf{1}_\Omega, h, u_h)$ and the widths above (${\approx}6.4$M parameters).}

\paragraph{Parameter counts.} The canonical 2D model totals ${\approx}11.2$M parameters: shape encoder ${\approx}3.1$M, kernel head ${\approx}1.7$M, field lift ${\approx}6.4$M.

\begin{revblock}
\paragraph{3D configuration (MCB-B).} The shape encoder uses $d_{\text{model}}=192$, $64$ slice tokens, $4$ transformer layers, $4$ heads, and Fourier features with $10$ bands; the kernel head uses $2$ cross-attention layers at $d_{\text{model}}=192$ with Fourier bands $10$ for $p$ and $\zeta$ and $4$ for the normal, and a soft $\tanh$ cap of the log-density at $\log K_{\max}=15$ (kernel total $2.77$M parameters). The 3D lift is a cross-attention head at $d_{\text{model}}=192$ with $3$ cross-attention layers, whose query is a Fourier embedding of $p$ and whose context is the frozen shape latent together with $256$ source tokens ($384$ for Fitting) produced by a slice aggregator over source samples $(q_j, f(q_j))$. Its output is multiplied by $\max(0, -\mathrm{SDF}(p))$, so it vanishes on $\partial\Omega$. It receives neither $h$ nor $u_h$.
\end{revblock}

\section{Loss specifications and training schedule}\label{app:losses}

This section gives the explicit forms of the four kernel-training loss terms ($\mathcal{L}_{\text{NLL}}, \mathcal{L}_{\text{MV}}, \mathcal{L}_{\text{BL}}, \mathcal{L}_Z$), the loss-weight schedule actually used to obtain the reported numbers, and the optimizer / learning-rate setup for both training stages\rev{, followed by the WoS supervision budget and the training cost}.

\subsection{Kernel losses (Stage 1)}\label{app:losses-kernel}

For each interior query point $p$, the network output \rev{is} $\widetilde K_\theta(p,\zeta;\Omega)$\rev{; during training it is kept close to unit mass by $\mathcal{L}_Z$ below, and at inference it is normalized over the boundary quadrature (\S\ref{sec:method-kernel})}. With a surface quadrature $\{(\zeta_i, w_i)\}_{i=1}^{N_s}$ on $\partial\Omega$, define
\begin{equation}
    \log Z(p) \;=\; \log\!\sum_{i=1}^{N_s} w_i \exp(\log\widetilde K_\theta(p,\zeta_i;\Omega)),
\end{equation}
the log of the kernel's surface mass.

\paragraph{$\mathcal{L}_{\text{NLL}}$ (WoS exit-point likelihood).} Walk-on-Spheres simulation provides a boundary exit point $\zeta^*_i$ for each interior anchor $p_i$. We supervise
\begin{equation}
    \mathcal{L}_{\text{NLL}} \;=\; -\frac{1}{|S_{\text{valid}}|}\sum_{i \in S_{\text{valid}}} \log K_\theta(p_i, \zeta^*_i;\Omega),
\end{equation}
averaged over WoS walks that reached the boundary inside the step budget; otherwise the entry is masked. \rev{In 3D, the kernel is trained with this loss and the three losses below. In 2D, the exit points of $10^4$ precomputed walks per probe are smoothed into a Gaussian KDE with bandwidth $\sigma$ evaluated at $512$ boundary nodes. At each step we sample $64$ of these nodes, normalize both the kernel and the target over them, and minimize the KL divergence from the target plus $0.5$ times the $L_1$ distance between the two densities; the 2D kernel uses no $\mathcal{L}_{\text{MV}}$, $\mathcal{L}_{\text{BL}}$, or $\mathcal{L}_Z$.}

\paragraph{$\mathcal{L}_{\text{MV}}$ (mean-value martingale).} Treating $K_\theta(\cdot, \zeta)$ as a (signed) function of the interior point, the mean-value property requires
\begin{equation}
    \mathcal{L}_{\text{MV}} \;=\; \mathbb{E}_{p, \zeta, r}\!\left[\Big(K_\theta(p, \zeta) - \tfrac{1}{S}\sum_{s=1}^{S} K_\theta(p + r\mathbf{d}_s, \zeta)\Big)^2\right],
\end{equation}
with $S$ sphere samples $\mathbf{d}_s$ drawn uniformly on $S^{d-1}$ and radius $r$ log-uniform in $[0.2,\,0.9]\cdot\mathrm{SDF}(p)$. Both $K_\theta(p,\zeta)$ and the $K_\theta(p+r\mathbf{d}_s,\zeta)$ are normalized internally via the surface quadrature so that the discrepancy compares densities on a common scale. Batch entries for which $\mathrm{SDF}(p)$ falls below $10^{-3}\cdot\mathrm{diag}(\mathrm{bbox})$ are masked out (sphere-degenerate regime). MCB experiments use $S = 32$.

\paragraph{$\mathcal{L}_{\text{BL}}$ (boundary-limit peak).} For each surface anchor $\zeta_0 \in \partial\Omega$ with outward unit normal $\nu_{\zeta_0}$, we place a probe point $p_\epsilon = \zeta_0 - \epsilon \nu_{\zeta_0}$ just inside $\Omega$, with $\epsilon$ adapted iteratively until $\mathrm{SDF}(p_\epsilon) < -\epsilon/2$. The harmonic measure $\omega_{p_\epsilon}$ should concentrate at $\zeta_0$, so we drive $K_\theta(p_\epsilon, \zeta_0)$ up while penalizing mass placed away from $\zeta_0$:
\begin{equation}
    \mathcal{L}_{\text{BL}} \;=\; -\log K_\theta(p_\epsilon, \zeta_0;\Omega) \;+\; \gamma \!\!\sum_{i:\,\|\zeta_i - \zeta_0\| > \delta} w_i\, K_\theta(p_\epsilon, \zeta_i;\Omega),
\end{equation}
with defaults $\epsilon = 0.02$, $\gamma = 1.0$, $\delta = 0.1$. The summation uses the same surface quadrature as $\log Z$ except for the entry at $\zeta_0$, which is excluded.

\paragraph{$\mathcal{L}_Z$ (soft mass normalization).} We penalize deviation of $\log Z(p)$ from $0$ via a Huber loss
\begin{equation}
    \mathcal{L}_Z \;=\; \mathrm{Huber}_\delta(\log Z(p)) \;=\; \begin{cases}(\log Z(p))^2 & |\log Z(p)| \leq \delta \\ 2\delta\,|\log Z(p)| - \delta^2 & |\log Z(p)| > \delta\end{cases}
\end{equation}
with $\delta = 1$. Huber prevents spikes in $\log\widetilde K_\theta$ from producing outsized updates (an instability we observed under a pure $\ell_2$ penalty).

\subsection{Loss-weight schedule}\label{app:losses-schedule}

The total kernel loss is $\mathcal{L}_K = \lambda_{\text{NLL}}\mathcal{L}_{\text{NLL}} + \lambda_{\text{MV}}\mathcal{L}_{\text{MV}} + \lambda_{\text{BL}}\mathcal{L}_{\text{BL}} + \lambda_Z\mathcal{L}_Z$. The weights are step-dependent:

\begin{center}
\begin{tabular}{lcccc}
\toprule
Step range & $\lambda_{\text{NLL}}$ & $\lambda_{\text{MV}}$ & $\lambda_{\text{BL}}$ & $\lambda_Z$ \\
\midrule
$[0,\,10\text{k})$ \emph{(warm-in)} & 1.0 & 0.1 & 0.5 & 1.0 \\
$[10\text{k},\,80\text{k})$ \emph{(main)} & 1.0 & 1.0 & 0.5 & 1.0 \\
$[80\text{k},\,\infty)$ \emph{(MV-emphasis)} & 0.5 & 2.0 & 0.5 & 1.0 \\
\bottomrule
\end{tabular}
\end{center}

The warm-in stage suppresses $\mathcal{L}_{\text{MV}}$ at initialization, where the spherical-average targets and the kernel at $p$ are both moving and the loss can dominate the NLL signal before either has a useful shape. MCB-B Stage 1 runs for $30{,}000$ steps total, so only the warm-in and main ranges are reached for the headline numbers in \S\ref{sec:exp-mcb}; the post-$80$k MV-emphasis branch is provided in code but is not used to obtain reported MCB results.

\subsection{Stage 1 optimizer and LR schedule}\label{app:losses-stage1}

AdamW with weight decay $0.01$ on linear weights (no decay on biases / norm parameters). Linear warmup over $T_w = 1000$ steps to $\mathrm{lr}_{\max} = 3\cdot 10^{-4}$, then cosine decay to $\mathrm{lr}_{\min} = 10^{-5}$ over total $T = 30{,}000$ steps; gradient clipping at global norm $1.0$. Per step, the kernel sees $N_s = 2000$ surface samples and $N_p = 512$ interior anchors, with $B_q = 8$ queries per shape and one shape per gradient step. The kernel head's MLP score is soft-clipped via $\tanh$ to $\log\widetilde K_\theta(p,\zeta;\Omega) \in [-\log K_{\max},\,\log K_{\max}]$ with $\log K_{\max} = 15$. \rev{These are the 3D settings. The 2D kernel is trained for $60{,}000$ steps with AdamW and a one-cycle cosine schedule (warmup $500$ steps, peak learning rate $3\cdot 10^{-4}$, final $10^{-5}$) and no logit cap.}

\subsection{Stage 2 optimizer and LR schedule, with warm-start}\label{app:losses-stage2}

With $K_\theta$ frozen, $v_\varphi$ is trained against the masked pixel-wise mean-squared error
\begin{equation}\label{eq:lift-loss}
\mathcal{L}_v \;=\; \frac{1}{|\Omega|}\sum_{p \in \Omega} \big(u_{\text{pred}}(p) - u_{\text{true}}(p)\big)^2,
\end{equation}
in $y$-normalized space, where $u_{\text{pred}} = u_h + v_\varphi$ via the frozen kernel and $|\Omega|$ counts interior pixels. We use AdamW (weight decay $0.01$, default betas) with the same linear-warmup-then-cosine schedule from \S\ref{app:losses-stage1} but $T_w = 200$ and a per-category total step count (typically $30{,}000$--$50{,}000$). Gradient clipping is unchanged. The lift consumes $N_p = 512$ source-probe samples per problem.

The schedule is extended by warm-start: we reload the previous-best lift weights and resume under a fresh cosine schedule (optimizer state and step counter reset). The hardest categories reach $\sim 100$k effective steps after one or two warm-start rounds.

\rev{In 2D, the field lift is trained for $10{,}000$ steps on the Poisson pairs. For the variant of \S\ref{sec:discussion}, the source lift is trained with the same loss for $30{,}000$ steps on Laplace and Poisson pairs, and the residual head is then trained for $10{,}000$ steps on $u_{\text{pred}} = u_h + v_\varphi + r$ with $K_\theta$ and $v_\varphi$ frozen, using AdamW with weight decay $0.01$ and a one-cycle cosine schedule (warmup $300$ steps, peak learning rate $3\cdot 10^{-4}$, final $10^{-5}$).}

\begin{revblock}
\subsection{WoS supervision budget and variance}\label{app:wos-budget}

Table~\ref{tab:wos-budget} lists the WoS settings used for the reported kernels. The sampler runs on the GPU and completes ${\sim}5\times 10^8$ walks per second on an A100 even at the stricter termination $\varepsilon = 10^{-4}$. The whole 2D supervision ($5{,}000$ shapes $\times\ 32$ probes $\times\ 10^4$ walks, precomputed once) therefore takes seconds of GPU time, and the online supervision of one 3D category ($30{,}000$ steps $\times\ 8$ probes $\times\ 4$ exits ${\approx}\,10^6$ walks) runs in under a second. Table~\ref{tab:wos-percat} reports the per-category throughput and walk length in 3D at the training settings. Masked walks are rare ($0.0095\%$ in 2D, $0.05$--$0.40\%$ per 3D category).

\begin{table}[h]
\centering
\small
\caption{WoS supervision settings of the reported kernels. Masked: fraction of walks that do not reach the $\varepsilon$-shell within the step cap.}
\label{tab:wos-budget}
\begin{tabular}{lcc}
\toprule
 & 2D MNIST & 3D MCB-B \\
\midrule
$\varepsilon$ (normalized domain) & $10^{-3}$ & $10^{-3}$ \\
step cap & $128$ & $128$ \\
walks per probe & $10^4$ (precomputed) & $4$ fresh exits per step (online) \\
probes & $32$ per shape & $8$ per gradient step \\
target & KDE, $\sigma = 0.2\%$ of domain width, $512$ nodes & exit-point likelihood \\
masked & $0.0095\%$ & $0.05$--$0.40\%$ (by category) \\
\bottomrule
\end{tabular}
\end{table}

\begin{table}[h]
\centering
\small
\caption{Per-category WoS statistics in 3D (A100): throughput and mean number of steps per walk.}
\label{tab:wos-percat}
\begin{tabular}{lccccc}
\toprule
 & Nut & Gear & Motor & Fitting & Screws \\
\midrule
walks per second & $7.6{\times}10^8$ & $8.1{\times}10^8$ & $7.5{\times}10^8$ & $7.8{\times}10^8$ & $7.8{\times}10^8$ \\
mean steps per walk & $14.0$ & $12.8$ & $15.9$ & $12.8$ & $14.3$ \\
masked fraction & $0.125\%$ & $0.045\%$ & $0.110\%$ & $0.076\%$ & $0.402\%$ \\
\bottomrule
\end{tabular}
\end{table}

\paragraph{Variance.} The standard deviation of the WoS estimate falls as $1/\sqrt{N}$ in the number of walks $N$: for $h = x$ it drops from $0.031$ at $N = 100$ to $0.003$ at $N = 10^4$. Kernel quality is stable above ${\sim}1{,}000$ walks per probe (Appendix~\ref{app:abl-wos}), and the error of the 2D KDE targets is unchanged with $100\times$ more walks (Appendix~\ref{app:residual-origin}). At the canonical $10^4$ walks, supervision noise is therefore well below the kernel-fit error.

\subsection{Training cost}\label{app:train-cost}

On a single A100, the 2D kernel trains in ${\sim}1$\,h ($60{,}000$ steps) and the 2D field lift adds ${\sim}1$\,h; the source lift and the residual head of the variant take about $8$\,h and $1.5$\,h. A 3D kernel takes $8.8$\,h (Nut) to $24$\,h (Motor, on a shared GPU) per category ($30{,}000$ steps).
\end{revblock}

\section{Baseline implementations}\label{app:baselines}

We use the authors' published source code wherever it is available. \textbf{Transolver}~\citep{wu2024transolver}, \textbf{LNO}~\rev{\citep{wang2024latent}}, and \textbf{UPT}~\citep{alkin2024universal} are run from the official public repositories of their respective papers; we adapt only the data loaders to our $(\Omega, h, f) \mapsto u$ format and otherwise keep architectures, optimizers, and training schedules at the published defaults. \rev{\textbf{BENO}~\citep{wang2024beno} is run from its official implementation with the data pipeline adapted to our format; we use $512$ boundary samples and train for $160$ epochs at $64^2$ followed by $40$ epochs at $128^2$, the resolution of all other methods.} For the 3D MCB-B Poisson benchmark, we additionally use the dataset and reference solutions released by NGF~\citep{yoo2025neural} on their public GitHub repository, which provides the FEM tetrahedral meshes and ground-truth solutions used in their Table~2; we evaluate on the same shape and $(h, f)$ test split, allowing direct head-to-head comparison without re-running their FEM pipeline. For the 2D MNIST benchmark, however, the NGF authors did not release a 2D code path or 2D evaluation data; we therefore \rev{ported their official 3D implementation to 2D (Appendix~\ref{app:ngf-2d})}. Numbers reported for 2D NGF reflect this \rev{port} rather than the authors' code.

\section{Intuitive demonstrations: details}\label{app:intuitive}

These are proof-of-concept demos used in \S\ref{sec:exp-intuitive}; the formal benchmarks of \S\ref{sec:exp-mnist} and \S\ref{sec:exp-mcb} train per-category as standard for those protocols, so the shared-kernel framing here is specific to these demos.

\subsection{3D harmonic on common graphics meshes}\label{app:intuitive-harmonic}

\paragraph{PDE and shapes.} Dirichlet Laplace, $\Delta u = 0$ in $\Omega$ with $u|_{\partial\Omega} = h$, on four unit-cube-normalized meshes: armadillo, bunny, fandisk, lucy. Boundary data $h \in \{\sin x,\,\sin z\}$, giving eight test cases in total.

\paragraph{Ground truth.} FEM solutions on volumetric tetrahedral meshes per shape. Final volumes are exported as $256^3$ voxel grids of the harmonic field for high-resolution rendering, with rel-$L_2$ measured against the FEM reference inside the FEM interior mask.

\paragraph{Ours.} A residual-distilled NHMO export. A single shape-conditioned harmonic kernel $K_\theta$ is fitted once across all four shapes, followed by a residual head trained against the FEM reference and distilled into a single forward pass for visualization-quality output.

\paragraph{Green-function-style baseline.} A learned volumetric Green's function as in prior work~\citep{yoo2025neural,boulle2022data,teng2022learning,negi2024learning,gin2021deepgreen,li2020multipole,teixeira2026variational}, evaluated against the same FEM reference on the same volumes.

\begin{table}[h]
\centering
\caption{3D harmonic on common graphics meshes: per-case rel-$L_2$ against FEM reference.}
\label{tab:intuitive_3d}
\begin{tabular}{llccc}
\toprule
shape & $h$ & Ours & GF style & ratio \\
\midrule
armadillo & $\sin x$ & 0.011 & 0.117 & $10.4{\times}$ \\
armadillo & $\sin z$ & 0.011 & 0.103 & $9.2{\times}$ \\
bunny     & $\sin x$ & 0.019 & 0.249 & $13.2{\times}$ \\
bunny     & $\sin z$ & 0.016 & 0.214 & $13.4{\times}$ \\
fandisk   & $\sin x$ & 0.011 & 0.142 & $12.6{\times}$ \\
fandisk   & $\sin z$ & 0.012 & 0.138 & $11.8{\times}$ \\
lucy      & $\sin x$ & 0.006 & 0.123 & $21.3{\times}$ \\
lucy      & $\sin z$ & 0.009 & 0.045 & $5.1{\times}$ \\
\midrule
mean & & \textbf{0.012} & 0.142 & $11.8{\times}$ \\
\bottomrule
\end{tabular}
\end{table}

\begin{figure}[h]
\centering
\includegraphics[width=\linewidth]{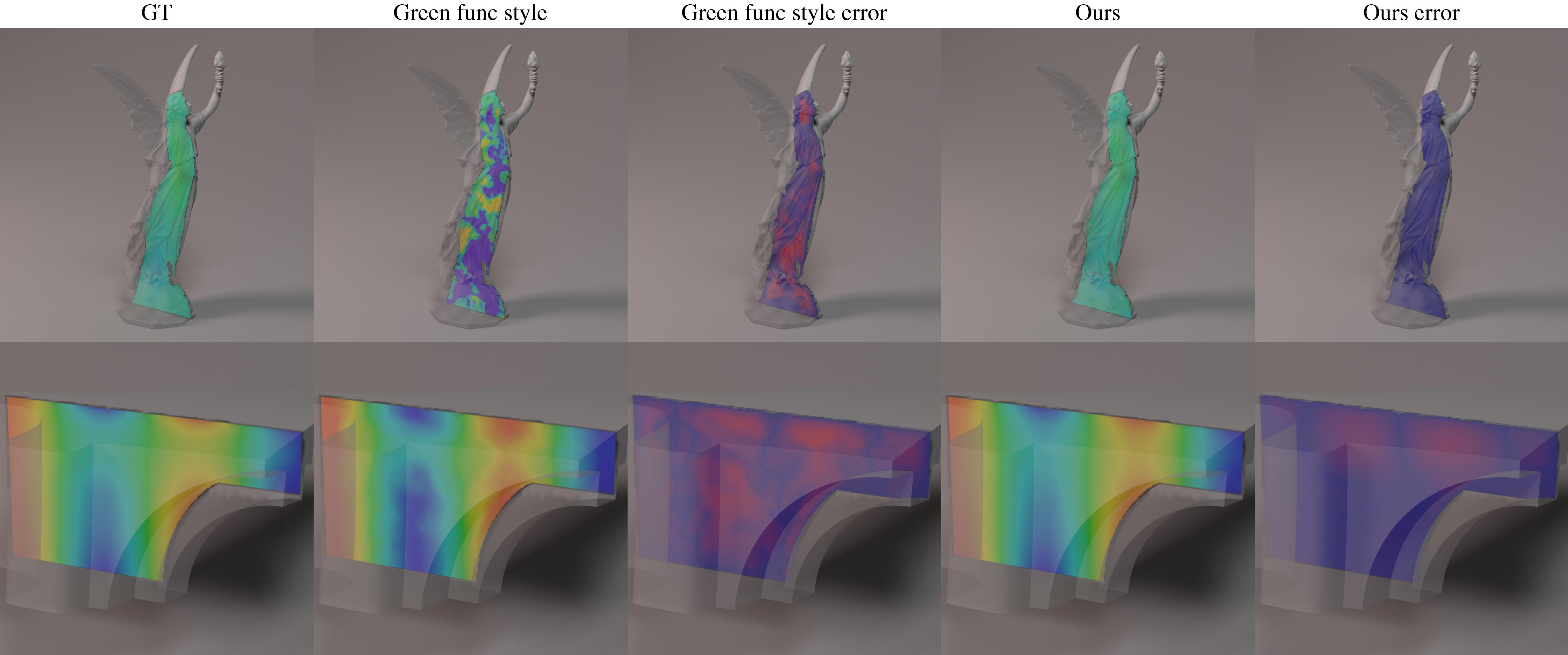}
\caption{Additional intuitive 3D harmonic comparisons. Top: lucy. Bottom: fandisk. Same five-column layout as Figure~\ref{fig:teaser}.}
\label{fig:teaser-appx}
\end{figure}

\subsection{Drift adaptation on a bunny slice}\label{app:intuitive-drift}

\paragraph{PDE.} Constant-drift Laplace,
\begin{equation}
\Delta u + \beta \cdot \nabla u \;=\; 0 \quad \text{in } \Omega, \qquad u|_{\partial\Omega} \;=\; h,
\end{equation}
on a 2D $y{=}0$ slice of a bunny mesh ($\Omega \subset \R^2$ is the slice interior). Dirichlet boundary data $h \in \{\sin x,\,\sin z\}$. Drift vectors $\beta$ are listed in Table~\ref{tab:drift_cases}.

\paragraph{Ground truth.} Computed by Walk-on-Spheres with a Yukawa-style transform that absorbs the drift as a path-dependent killing factor.

\paragraph{Ours.} Warm-started from the same frozen $K_\theta$ used in \S\ref{app:intuitive-harmonic}. We attach a small drift-conditioned adapter and a residual head; the kernel itself is not retrained.

\paragraph{Green-function-style baseline.} The same construction as in \S\ref{app:intuitive-harmonic}, also warm-started from the frozen Laplace kernel and conditioned on the drift parameter, but without the residual head.

\paragraph{Training budget.} Both methods are trained under identical settings, namely 4000 optimization steps, batch size 128, a single learning rate, and an 80/20 pixel split over eight $256{\times}256$ slice cases (four drift vectors $\times$ two boundary signals). Training samples are pixels rather than fixed epochs; we therefore report this as a matched optimization-budget comparison.

\begin{table}[h]
\centering
\caption{Drift adaptation: per-slice test rel-$L_2$ on the bunny slice.}
\label{tab:drift_cases}
\begin{tabular}{llcc}
\toprule
drift $\beta$ & boundary $h$ & Ours & GF style \\
\midrule
$(2,0,0)$    & $\sin x$ & 0.060 & 0.366 \\
$(2,0,0)$    & $\sin z$ & 0.062 & 0.280 \\
$(-2,0,0)$   & $\sin x$ & 0.062 & 0.378 \\
$(-2,0,0)$   & $\sin z$ & 0.065 & 0.326 \\
$(0,0,2)$    & $\sin x$ & 0.069 & 0.364 \\
$(0,0,2)$    & $\sin z$ & 0.061 & 0.293 \\
$(1.5,1,0)$  & $\sin x$ & 0.060 & 0.360 \\
$(1.5,1,0)$  & $\sin z$ & 0.060 & 0.288 \\
\midrule
mean & & \textbf{0.062} & 0.323 \\
\bottomrule
\end{tabular}
\end{table}

\begin{figure}[h]
\centering
\includegraphics[width=\linewidth]{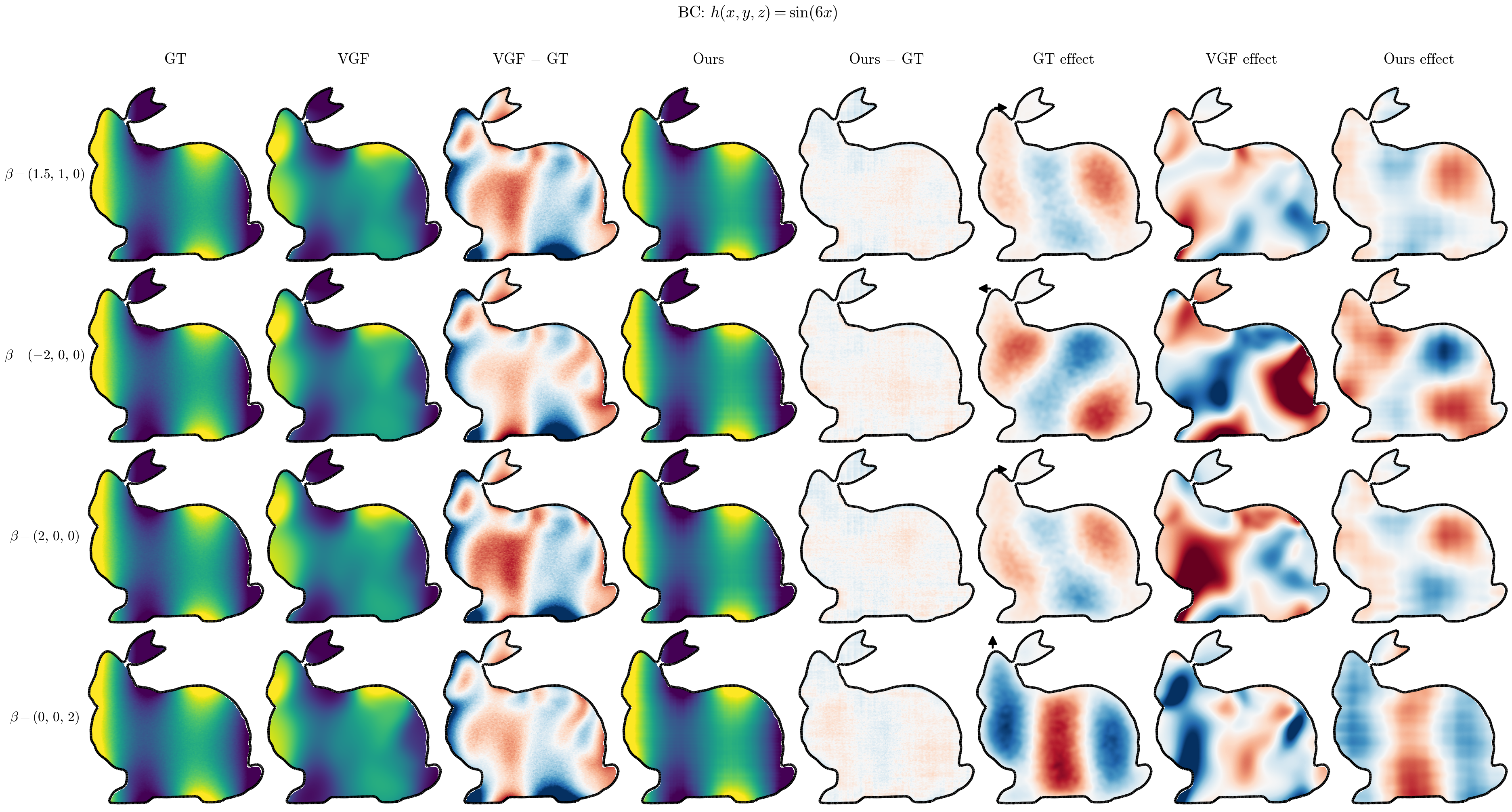}
\caption{Bunny drift-diffusion qualitative, $h(x,y,z) = \sin(6x)$. Rows: four drift vectors $\beta = (1.5, 1, 0)$, $(-2, 0, 0)$, $(2, 0, 0)$, $(0, 0, 2)$. Columns 1--5: GT, GF style, GF style $-$ GT, Ours, Ours $-$ GT. Columns 6--8: drift-induced field $u_\beta$ minus the mean over the four drifts, highlighting the dipole structure aligned with each $\beta$ (Gaussian-blurred for clarity; metrics in Table~\ref{tab:drift_cases} use unblurred fields).}
\label{fig:drift-sinx}
\end{figure}

\begin{figure}[h]
\centering
\includegraphics[width=\linewidth]{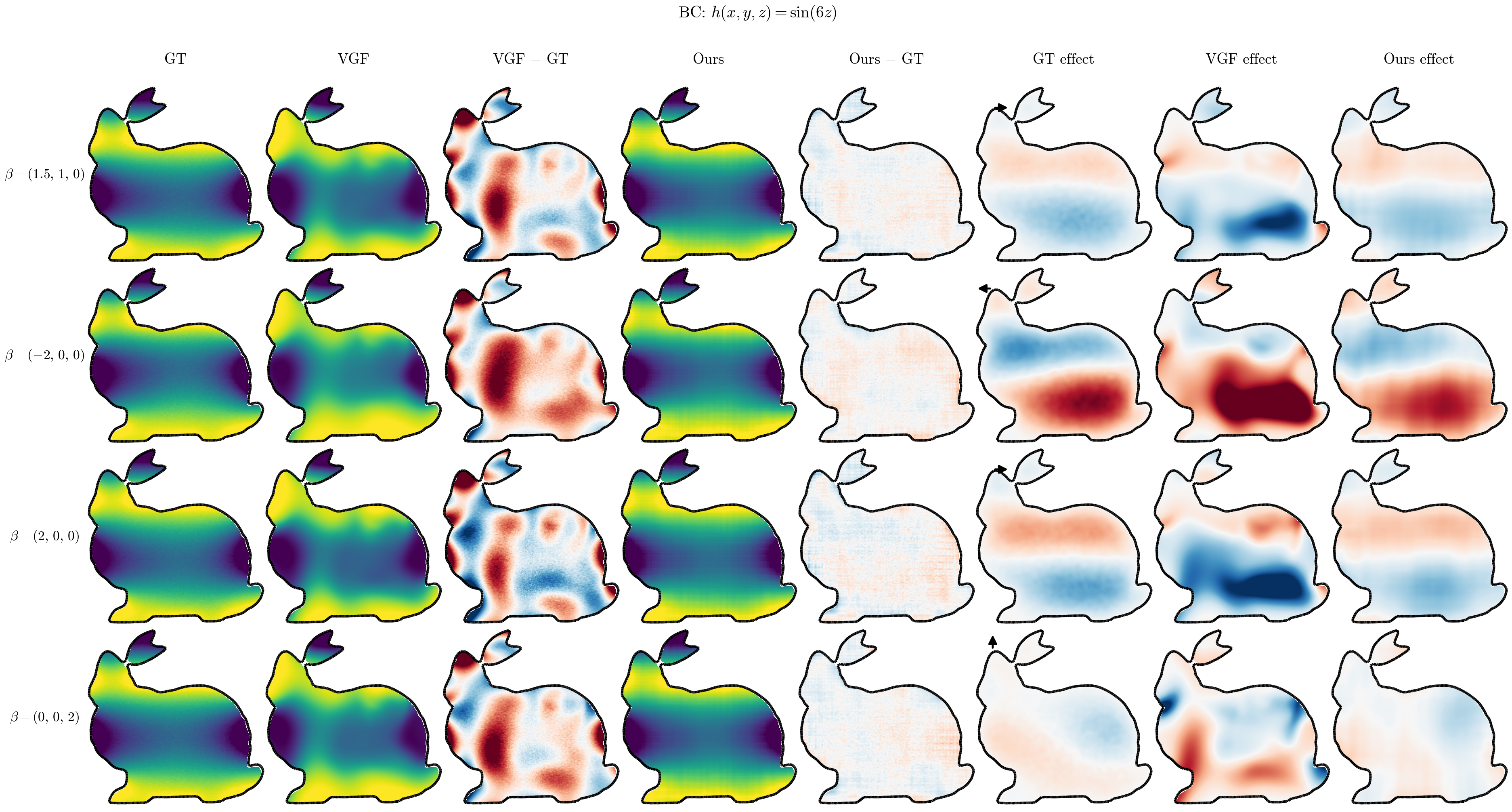}
\caption{Bunny drift-diffusion qualitative, $h(x,y,z) = \sin(6z)$. Same layout as Figure~\ref{fig:drift-sinx}.}
\label{fig:drift-sinz}
\end{figure}

\section{2D MNIST benchmark: setup, \rev{NGF port}, and additional qualitative}\label{app:mnist}

\subsection{Setup details}\label{app:mnist-setup}

\paragraph{Geometry.} Each shape is an MNIST digit raster upsampled from $28{\times}28$ to a $256{\times}256$ binary mask, optionally retaining the thin inner holes that arise from the digit topology. The interior mask, ${\sim}512$ boundary samples with normals, and interior anchors are produced by a deterministic shape generator. Domains for digits 0, 6, 8, 9 are multiply-connected.

\paragraph{Boundary-condition families.} Two parametric families are sampled per problem with random coefficients,
\begin{align*}
\texttt{poly3:}\quad   & h(x,y) = a(x^3 - 3xy^2) + b(y^3 - 3x^2 y) + c\,x^2 \\
\texttt{exp\_mix:}\quad & h(x,y) = a\, e^{0.5x}\cos(0.5y) + b\, x\, y^2 + c\, y
\end{align*}
In-distribution coefficients $a, b, c \sim U[-1, +1]$; OOD coefficients $a, b, c \sim U[+1, +2]$, strictly outside training. Two earlier high-frequency families \texttt{trig1}, \texttt{trig2} are kept for ablation only and excluded from headline numbers because every learned method failed catastrophically on them.

\paragraph{Sources.} Poisson problems use one of four source families: \texttt{sin\_cos}, \texttt{polynomial}, \texttt{gaussian}, \texttt{asymmetric}.

\paragraph{Splits.} 991 train / 50 test (in-dist) / 50 OOD shapes; 7500 / 408 / 397 problem instances after filtering.

\paragraph{Resolution.} Numerical ground truth is a 5-point finite-difference Poisson solver at $256^2$ followed by bilinear downsampling to $128^2$, the resolution at which all neural models train and evaluate.

\paragraph{Eval metric.} Un-normalized relative-$L_2$ error over interior pixels of each shape ($\text{mask}=1$), aggregated across all problem instances per split. Trainer-side losses on $y$-normalized residuals are not used.

\begin{revblock}
\subsection{NGF 2D port}\label{app:ngf-2d}

NGF's official code is written for tetrahedral meshes. Its network sees only positional encodings of the vertex coordinates, so its per-point features depend only on the geometry; three linear heads $A$ (interior), $C$ (all points), and $D$ (boundary) produce the interior solution as $A(C^\top \mathrm{rhs}) - A(D^\top h)$ up to a diagonal scaling, where in the released MCB-B configuration a learned mass head forms $\mathrm{rhs}$ from the source. The boundary values are given, not predicted. Our 2D port keeps this architecture and the official optimization settings (feature width $128$, learning rate $10^{-4}$, gradient clipping $0.5$, effective batch $8$) and makes the adaptations a pixel grid requires: the encoder receives the interior mask (the pixel lattice is identical across shapes, so geometry must enter through the mask), the boundary is the band of exterior pixels adjacent to the domain, and multiply-connected boundaries are handled by that band without change.

An initial port differed from the official setup in several respects: it omitted the mass head; it used feature width $64$, an MSE loss, and learning rate $5\cdot 10^{-4}$; it regressed $y$-normalized targets (the NGF forward pass is linear in the data and has no bias path, so it cannot represent the offset this normalization introduces); it evaluated the boundary term on $256$ randomly subsampled band pixels per step; and its encoder also received $h$ and $f$. The aligned port follows the official setup in all of these respects, and Table~\ref{tab:ngf-port} compares the two.

\begin{table}[h]
\centering
\footnotesize
\caption{NGF 2D port: initial port versus the port aligned with the official setup (relative $L_2$, \%, mixed splits).}
\label{tab:ngf-port}
\begin{tabular}{lcccc}
\toprule
 & test mean / median & test p95 / max & OOD mean / median & OOD p95 / max \\
\midrule
initial & $41.8$ / $22.1$ & --- & $24.5$ / $18.1$ & --- \\
aligned (40 epochs) & $3.87$ / $2.00$ & $16.9$ / $40.2$ & $4.20$ / $3.85$ & $6.0$ / $10.1$ \\
\bottomrule
\end{tabular}
\end{table}
\end{revblock}

\subsection{Additional qualitative comparisons}\label{app:mnist-extra}

\begin{figure}[h]
    \centering
    \includegraphics[width=\linewidth]{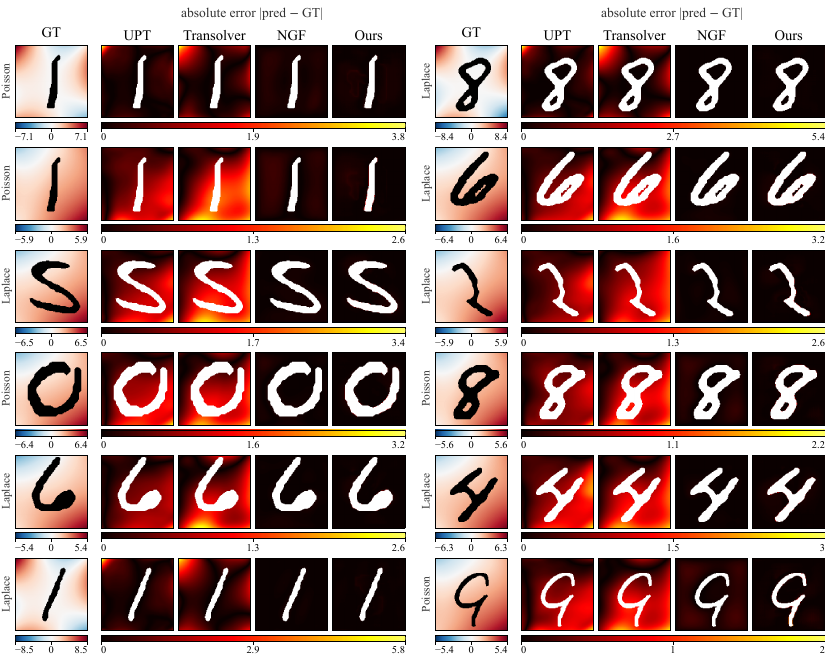}
    \caption{2D MNIST OOD qualitative (additional, set 1 of 2). $12$ shapes, random pick from remaining Laplace and Poisson cases\rev{; each GT tile is tagged with its problem type}.}
    \label{fig:mnist-extra-1}
\end{figure}

\begin{figure}[h]
    \centering
    \includegraphics[width=\linewidth]{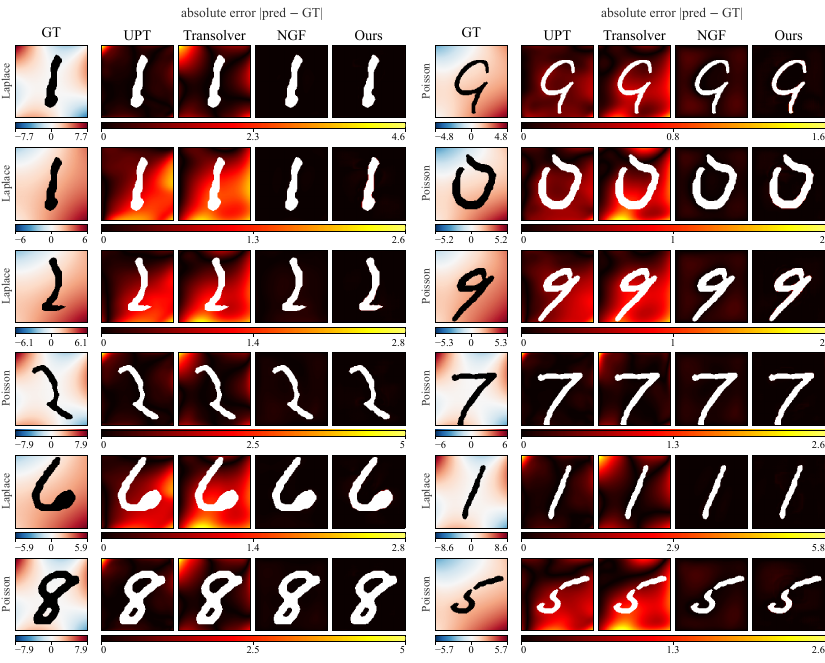}
    \caption{2D MNIST OOD qualitative (additional, set 2 of 2). $12$ shapes, random pick from remaining Laplace and Poisson cases\rev{; each GT tile is tagged with its problem type}.}
    \label{fig:mnist-extra-2}
\end{figure}

Figures~\ref{fig:mnist-extra-1} and~\ref{fig:mnist-extra-2} extend Figure~\ref{fig:mnist_qualitative} with $24$ additional OOD shapes (random pick from remaining Laplace and Poisson cases), same per-row layout \rev{and color-scale convention}.

\section{Additional MCB-B qualitative comparisons}\label{app:fig4-extra}

\begin{figure}[h]
    \centering
    \includegraphics[width=\linewidth]{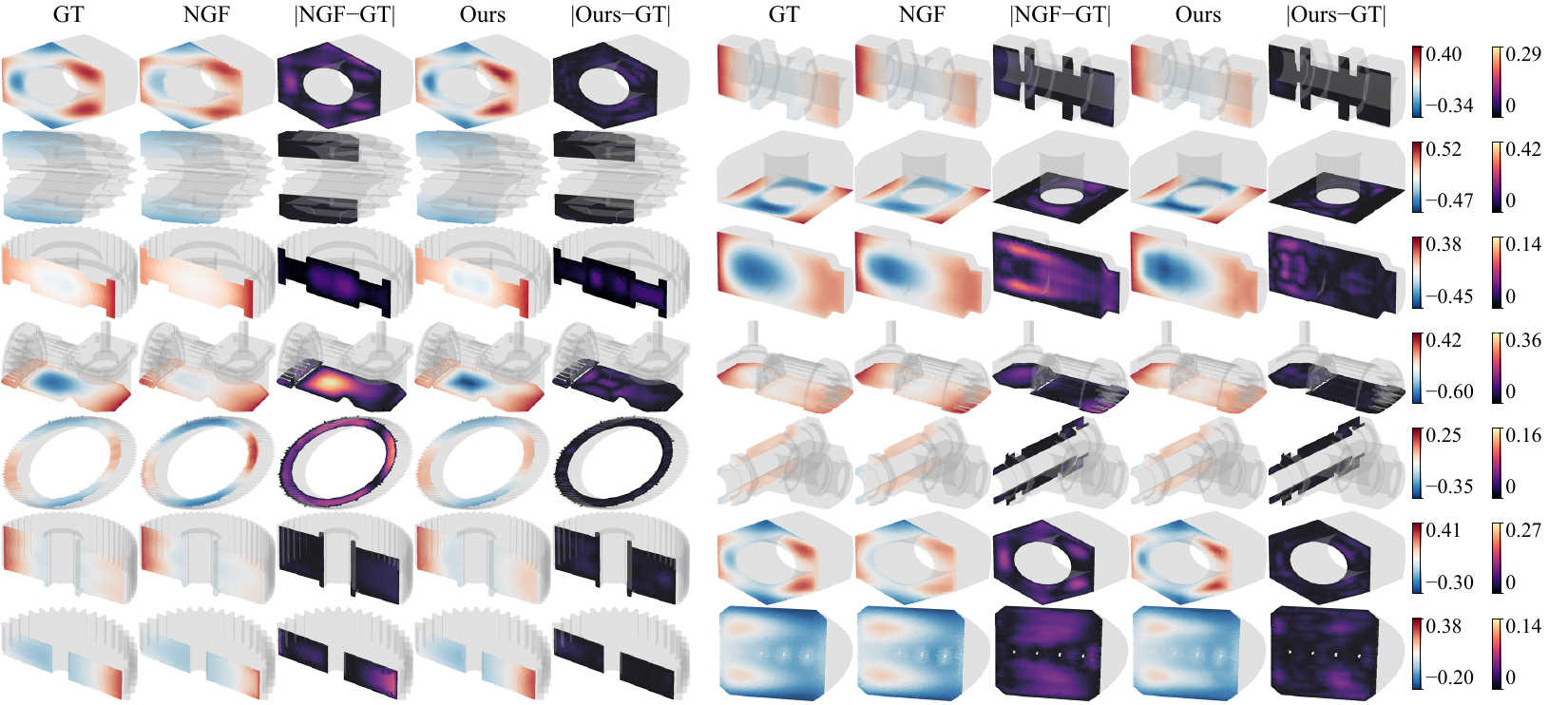}
    \caption{\textbf{Additional MCB-B Poisson qualitative comparisons.} 14 shapes (7 rows $\times$ 2 shapes per row) in the same layout \rev{and color-scale convention} as Figure~\ref{fig:qualitative}.}
    \label{fig:qualitative-appx}
\end{figure}

Figure~\ref{fig:qualitative-appx} extends the qualitative comparison of \S\ref{sec:exp-mcb} with 14 additional shapes, in the same per-shape five-panel layout (GT, NGF, $|\text{NGF}-\text{GT}|$, Ours, $|\text{Ours}-\text{GT}|$) and the same 3D cross-section rendering style.

\begin{revblock}
\section{3D coefficient-OOD study}\label{app:ood3d}

MCB-B's test problems use the same coefficient ranges as training. To test extrapolation in 3D without retraining, we drew $10$ test shapes per category and posed $4$ problems on each whose boundary data and sources come from parametric families with coefficients in $U[1,2]$, outside the training ranges. References are computed with the FEM solver (lapy) used in the NGF repository. NGF is run from its released mass-prediction checkpoints; ours is the canonical pipeline of Table~\ref{tab:mcb_main}. A second, Laplace-only track uses the same shifted boundary data with $f \equiv 0$ and isolates boundary extrapolation. Table~\ref{tab:ood3d} reports the results; the in-distribution reference for each method is its Table~\ref{tab:mcb_main} macro-average ($0.241$ for NGF, $0.193$ for ours).

\begin{table}[h]
\centering
\small
\caption{3D coefficient-OOD study: mean relative $L_2$ error over $40$ problems per category (identical problems for both methods).}
\label{tab:ood3d}
\begin{tabular}{llcccccc}
\toprule
Track & Method & Nut & Gear & Motor & Fitting & Screws & Macro \\
\midrule
Poisson-OOD & NGF (released) & 0.678 & 0.605 & 0.616 & 0.627 & 0.548 & 0.615 \\
            & Ours           & 0.382 & 0.099 & 0.347 & 0.211 & 0.274 & 0.263 \\
\midrule
Laplace-OOD & NGF (released) & 0.633 & 0.604 & 0.631 & 0.637 & 0.603 & 0.621 \\
            & Ours           & 0.127 & 0.037 & 0.162 & 0.070 & 0.101 & 0.099 \\
\bottomrule
\end{tabular}
\end{table}
\end{revblock}

\section{Inference speed: caching is implied by the factorization}\label{app:speed}

\paragraph{Bit-identity of the cache.} The kernel matrix $K_{\text{eff}} = [w_j\, K_\theta(p_i,\zeta_j;\Omega)]_{ij}$ is a deterministic function of $\Omega$ alone, so reusing it across $(h, f)$ on the same shape is fp32-bit-identical to recomputing it per problem; we verified this on $50$ random $(p, h)$ pairs. Memory: an $(n_{\text{in}} \times n_{\text{surf}})$ fp32 tensor, ${\approx}8$\,MB per shape at $128{\times}128$ with $200$ surface samples.

\begin{revblock}
\paragraph{Inference-only optimizations.} The 3D timings in \S\ref{sec:perf} use the following optimizations, which do not retrain or change any model. (i) \emph{Query folding} (per-problem solve): without folding, the 3D lift treats each query point as a separate batch entry that cross-attends to its own copy of the same context, recomputing the context keys and values per query; since the cross-attention block processes query tokens independently, we fold all queries into the sequence dimension and compute the keys and values once. Predictions are identical in fp32, and the relative $L_2$ errors against the FEM references are unchanged. (ii) \emph{Faster build}: the signed distance grid is rasterized on the GPU with exact point-triangle distances and a ray-parity inside test, which matches the CPU reference at all but isolated grid points, and $K_{\text{eff}}$ is evaluated with the quadrature normalization folded in and in bf16. The build also redraws its random surface and interior samples, so its predictions are not bitwise identical to the original pipeline; the change in mean relative $L_2$ error ($-0.004$ on Nut, $-0.002$ on Motor) is within that of a control that only redraws the samples ($-0.003$ and $+0.000$).
\end{revblock}

\paragraph{Why the parametric baselines cannot cache.} Transolver fuses $(\text{mask}, h, f)$ tokens through slice-attention where every layer mixes geometry and boundary data; UPT concatenates $(\text{mask}, h, f)$ as input channels to its image encoder\rev{; LNO and BENO likewise take the boundary data and the source as network inputs. None of these architectures separates a geometry-only state from $(h, f)$, so they admit no per-shape cache. NGF is different: its per-point features depend only on the geometry, so a similar split into a per-shape state and a per-problem read-out is possible for it in principle; in our timing, we preloaded its mesh on the GPU and changed only the boundary data (\S\ref{sec:perf}).} The cache is enabled by NHMO's factorization, not by an engineering choice.

\paragraph{Complexity.} Per-shape precompute is $O(n_{\text{in}}\, n_{\text{surf}}\, d_{\text{kernel}})$. Per-problem cost is $O((n_{\text{in}} + n_{\text{surf}})\,d) + O(R^2\, c\, D)$ for the lift U-Net at resolution $R$, base channels $c$, depth $D$. This matches the complexity class of the parametric baselines' Galerkin-style forward.

\paragraph{Caveats.} (i) When every problem uses a different shape ($K{=}1$), NHMO pays its \rev{geometry step (Table~\ref{tab:runtime}) for every problem; on a new mesh-based shape this step is cheaper than meshing, whereas on grid inputs the grid-based baselines need no such step}. (ii) Training is a separate concern\rev{ (Appendix~\ref{app:train-cost})}. The speed advantage is at deployment, where the system is queried many times against the same geometries.

\section{Ablations: detail}\label{app:ablations}

This appendix expands the five takeaways summarized in \S\ref{sec:ablation}. All numbers are un-normalized rel-$L_2$ over interior pixels at $128\times 128$, mixed (Laplace + Poisson) test split unless noted otherwise.

\begin{revblock}
\subsection{Separating the lift's roles: source lift and residual head}\label{app:ladder}

Table~\ref{tab:ladder} builds the 2D model up from the kernel alone. The source lift, which sees only $(\Omega, f)$, lowers the test error and degrades by only $1.04{\times}$ (mean) under the OOD shift, since $h$ enters it only through the linear boundary integral. Adding the residual head, trained on top of the frozen source lift, gives the three-term model $u = \langle h, K_\theta\rangle + v_\varphi(\Omega, f) + r(\Omega, h, u_h)$ of \S\ref{sec:discussion}. It matches or slightly surpasses the single $h$-conditioned lift of the main model at a larger total capacity, while keeping the source channel strictly independent of $h$. Training the lifts with five seeds (seed $0$ is the reported checkpoint; $\pm$ is the standard deviation over seeds) gives a test mean of $1.87 \pm 0.05\%$ for the three-term model and $2.09 \pm 0.03\%$ for the single lift of the main model (OOD $2.48 \pm 0.05\%$ and $2.60 \pm 0.05\%$); the three-term model is lower on test for every seed. The source lift alone is nearly seed-independent (test mean $6.09$--$6.10\%$), as expected if its error is dominated by the kernel-fit residual it cannot see.

\begin{table}[h]
\centering
\small
\caption{From the kernel alone to the three-term variant (relative $L_2$, \%, mean / median; Table~\ref{tab:mnist_main} scale).}
\label{tab:ladder}
\begin{tabular}{llcc}
\toprule
Variant & head inputs & test & test\_ood \\
\midrule
kernel only & --- & $7.7$ / $5.4$ & $6.8$ / $5.6$ \\
+ source lift $v_\varphi(\Omega, f)$ & $\mathbf{1}_\Omega$, SDF, $f$ & $6.1$ / $4.5$ & $6.3$ / $5.2$ \\
+ residual head $r(\Omega, h, u_h)$ & $\mathbf{1}_\Omega$, $h$, $u_h$ & $\mathbf{1.84}$ / $\mathbf{1.74}$ & $\mathbf{2.56}$ / $\mathbf{2.39}$ \\
\midrule
main model: single lift $v_\varphi(\Omega, h, f)$ & $\mathbf{1}_\Omega$, $h$, $f$, $u_h$ & $2.1$ / $2.0$ & $2.6$ / $2.5$ \\
\bottomrule
\end{tabular}
\end{table}

\paragraph{What the correction learns.} On Laplace pairs ($f \equiv 0$, $205$ test pairs) the source contribution is zero, so the output of the single $h$-conditioned lift of the main model there is exactly its boundary correction. We checked three possibilities. It is not random: it correlates with the kernel-fit residual $e_h = u - u_h$ at median $0.97$ and removes $71\%$ of it. It is not a fluctuation around the residual: $82\%$ of its spectral energy lies in the lowest tenth of radial frequencies, against $1\%$ for a matched white-noise control (medians), and it reproduces the residual rather than scattering around it. It is not a fixed bias: the residual it tracks is linear in $h$, changes sign and shape with the boundary data, and averages to about zero over the symmetric coefficient draw of the test split. Its magnitude is small (median $4.8\%$ of $\|u_h\|$), and removing the boundary inputs forfeits the correction (the source lift alone reaches $6.1\%$ test mean against $2.1\%$, Table~\ref{tab:ladder}). The residual head $r$ of the three-term model behaves the same way on these pairs (median correlation $0.97$, $75\%$ of the residual removed, $81\%$ low-frequency energy).

\paragraph{Perturbations of the boundary data.} The kernel channel is linear in $h$: a perturbation $h \to h + \epsilon\eta$ changes $u_h$ by exactly $\epsilon\langle \eta, K_\theta\rangle$, which is bounded by $\epsilon \max|\eta|$ for the quadrature-normalized kernel. We measured the amplification, the relative $L_2$ response of $u_h$ divided by $\epsilon\max|\eta|$, on $20$ shapes for $\epsilon \in [0.01, 0.5]$: its mean over shapes is $0.84$ for smooth $\eta$ and $0.17$ for white-noise $\eta$, constant over this range of $\epsilon$, and the full model including the learned lift stays at or below $1.02$ on average.
\end{revblock}

\subsection{Lift removal (kernel-only vs.\ kernel + lift)}\label{app:abl-lift-removal}

Defends the factorization $u = \langle h, K_\theta\rangle + v_\varphi$. The kernel alone already beats every \rev{nonlinear end-to-end} baseline on OOD; the learned lift is a small correction.

\begin{table}[h]
\centering
\caption{Lift removal. Mixed (Laplace + Poisson) rel-$L_2$ over interior pixels (median / mean).}
\label{tab:abl-lift-removal}
\begin{tabular}{lccc}
\toprule
Variant & test (in-dist) & test\_ood & OOD/test \\
\midrule
$K$-only (no $v_\varphi$)              & $5.4\%$ / $7.7\%$    & $5.6\%$ / $6.8\%$   & $\mathbf{1.0{\times}}$ \\
\textbf{$K + 6.4$M lift (canonical)}   & $\mathbf{2.0\%}$ / $\mathbf{2.1\%}$ & $\mathbf{2.5\%}$ / $\mathbf{2.6\%}$ & $1.25{\times}$ \\
$K + 11.3$M lift (capacity scan, A2)   & $2.0\%$ / $2.1\%$    & $2.3\%$ / $2.5\%$   & $1.15{\times}$ \\
\bottomrule
\end{tabular}
\end{table}

\subsection{Lift capacity ($6.4$M vs $11.3$M parameters)}\label{app:abl-capacity}

Doubling lift parameters from $6.4$M to $11.3$M gives no in-distribution improvement and only marginal OOD gain (Table~\ref{tab:abl-lift-removal}, last row). NHMO is not capacity-limited at the lift; the factorization, not network size, is the structural reason for the result.

\subsection{KDE bandwidth $\sigma$ for Walk-on-Spheres supervision}\label{app:abl-kde-sigma}

The kernel is robust to KDE $\sigma$ over a $5\times$ range ($0.1\%$--$0.5\%$ of domain width).

\begin{table}[h]
\centering
\caption{KDE bandwidth scan. $K$-only test median.}
\label{tab:abl-kde-sigma}
\begin{tabular}{lc}
\toprule
$\sigma$ (fraction of domain width) & $K$-only test median \\
\midrule
$0.1\%$ (sharper)                              & ${\sim}6.5\%$  \\
\textbf{$0.2\%$} (canonical)                   & $\mathbf{5.4\%}$ \\
$0.5\%$ (smoother)                             & ${\sim}6.0\%$  \\
\bottomrule
\end{tabular}
\end{table}

\subsection{Training-shape count and single mixed-corpus generalization}\label{app:abl-corpus}

NHMO is trained on a fixed $5{,}000$-shape MNIST corpus drawn from \emph{all} 10 digit classes ($0$--$9$), spanning both simply-connected (e.g., $1$, $7$) and multiply-connected (e.g., $0$, $6$, $8$, $9$) topologies, with no class labels. The harmonic-measure factorization makes the kernel a per-shape function of geometry, so corpus diversity adds signal \rev{rather than competing for capacity}. \rev{By contrast, NGF's published MCB-B numbers come from five separate models, one per shape category.} NHMO trains one kernel and one lift across all $10$ MNIST digit classes simultaneously.

\begin{table}[h]
\centering
\caption{Training-shape count. $K$-only test median rel-$L_2$ as the corpus grows.}
\label{tab:abl-corpus}
\begin{tabular}{lcc}
\toprule
Train shapes & $K$-only test median & $K + $lift test median \\
\midrule
$200$    & ${\sim}7.0\%$  & ---           \\
$500$    & ${\sim}6.0\%$  & ${\sim}3.5\%$ \\
$1{,}000$  & ${\sim}5.7\%$  & ${\sim}2.5\%$ \\
\textbf{$5{,}000$} (canonical) & $\mathbf{5.4\%}$ & $\mathbf{2.0\%}$ \\
\bottomrule
\end{tabular}
\end{table}

\subsection{Walk-on-Spheres sample count}\label{app:abl-wos}

The kernel is robust to the WoS sample count above $1{,}000$ walks per query; the canonical run uses $10{,}000$ walks per query and matches the test median of the kernel ablation in Table~\ref{tab:abl-lift-removal}. Below $1{,}000$ walks the KDE supervision becomes too noisy and the kernel degrades.

\subsection{Boundary \rev{quadrature resolution} at inference ($n_{\text{surf}}$ scan)}\label{app:abl-nsurf}

NHMO's boundary-integral $\sum_\zeta K_\theta(p,\zeta)\, h(\zeta)$ is the discretization of a continuous integral. We verify this at inference time with the same trained kernel, varying only the number of boundary samples $n_{\text{surf}}$. Above ${\sim}100$ samples the prediction is converged; below that, the result degrades \emph{gracefully} rather than catastrophically\rev{, so the model is robust to the boundary-quadrature resolution at inference over the tested range}. Parametric baselines have no analogous discretization knob; their inference quality is tied to whatever resolution the encoder was trained at.

\begin{table}[h]
\centering
\caption{Boundary-discretization scan at inference. Mixed rel-$L_2$ on test\_ood, $25$ shapes $\times$ ${\sim}8$ problems.}
\label{tab:abl-nsurf}
\begin{tabular}{lccc}
\toprule
$n_{\text{surf}}$ & median & mean & p$95$ \\
\midrule
$50$                          & $3.12\%$ & $3.58\%$ & $5.62\%$ \\
$100$                         & $2.48\%$ & $2.65\%$ & $4.05\%$ \\
\textbf{$200$} (canonical)    & $\mathbf{2.44\%}$ & $\mathbf{2.58\%}$ & $\mathbf{4.08\%}$ \\
$400$                         & $2.43\%$ & $2.56\%$ & $3.98\%$ \\
\bottomrule
\end{tabular}
\end{table}

\subsection{Per-MNIST-class breakdown (single mixed-corpus uniformity)}\label{app:abl-perclass}

A direct counter to per-category-corpora training. A single mixed-corpus NHMO model on the $5{,}000$-shape corpus (digits $0$--$9$) produces uniform performance across all classes; the spread across classes is much smaller than the \rev{OOD} gap to any \rev{nonlinear end-to-end} baseline.

\begin{table}[h]
\centering
\caption{Per-MNIST-class rel-$L_2$ mean (mixed Laplace + Poisson, full-interior un-normalized).}
\label{tab:abl-perclass}
\begin{tabular}{lccc}
\toprule
Digit class & $n_{\text{test}}$ & test mean & test\_ood mean \\
\midrule
0 & 50 & $2.14\%$ & $2.89\%$ \\
1 & 30 & $2.19\%$ & $2.95\%$ \\
2 & 40 & $1.86\%$ & $2.60\%$ \\
3 & 36 & $1.94\%$ & $2.26\%$ \\
4 & 38 & $2.15\%$ & $2.48\%$ \\
5 & 46 & $2.02\%$ & $2.48\%$ \\
6 & 35 & $2.28\%$ & $2.88\%$ \\
7 & 55 & $1.85\%$ & $1.95\%$ \\
8 & 39 & $2.38\%$ & $\mathbf{3.81\%}$ \\
9 & 39 & $2.32\%$ & $2.74\%$ \\
\midrule
spread (max $-$ min) & & $0.53\%$ & $1.87\%$ \\
\bottomrule
\end{tabular}
\end{table}

The $10$ digit classes have very different geometries ($1$ is a narrow stroke, $0$/$6$/$8$/$9$ have interior loops, $8$ has two), yet a single mixed-corpus model attains rel-$L_2$ within $0.53\%$ absolute spread on in-distribution test and $1.87\%$ on OOD. Digit $8$, the most challenging case (multiply-connected with two interior loops), is the worst class on OOD at $3.81\%$ but still beats every \rev{nonlinear end-to-end} baseline's overall mean.

\subsection{Representation invariance: SDF vs.\ point-cloud encoder}\label{app:abl-encoder}

A direct attack on the "your kernel just memorizes the boundary point cloud" critique. We retrain the kernel from scratch with the same hyperparameters as the canonical model ($d = 256$, $5{,}000$-shape corpus, KDE $\sigma = 0.2\%$, $60{,}000$ steps), changing \emph{only} the shape encoder family from a point-cloud encoder over $\partial\Omega$ to a 2D SDF-CNN over a $64\times 64$ SDF grid. The resulting kernel is paired with the canonical lift $v_\varphi$ (no lift retraining).

\begin{table}[h]
\centering
\caption{Representation-invariance ablation: identical hyperparameters, swap shape encoder.}
\label{tab:abl-encoder}
\begin{tabular}{lccc}
\toprule
Shape encoder & test (in-dist) & test\_ood & OOD/test \\
\midrule
\textbf{Point-cloud} (canonical)       & $\mathbf{2.0\%}$ / $\mathbf{2.1\%}$    & $\mathbf{2.5\%}$ / $\mathbf{2.6\%}$ & $1.25{\times}$ \\
$2$D SDF-CNN ($64\times 64$)            & $2.28\%$ / $2.43\%$ & $2.59\%$ / $2.97\%$ & $\mathbf{1.14{\times}}$ \\
\bottomrule
\end{tabular}
\end{table}

The encoder swap costs only $0.27\%$ absolute median on in-distribution test ($1.14\times$ canonical) and $0.09\%$ on OOD ($1.04\times$ canonical). Notably, the SDF-encoder kernel has a tighter OOD/test gap ($1.14\times$ vs $1.25\times$), suggesting the SDF representation may even improve coefficient-distribution generalization. Whatever representation makes $K_\theta(\cdot,\cdot;\Omega)$ an honest harmonic-measure operator suffices. \rev{NGF's released pipeline takes tetrahedral-mesh vertices with explicit boundary indices as input, so the same swap does not apply to it directly.}

\begin{revblock}
\subsection{Learned volumetric integrand}\label{app:gvol}

To test the Green's-function alternative of \S\ref{sec:why-not-green} directly, we replaced the source lift with a learned integrand $G_\theta(p, q; \Omega)$ whose volume integral against $f$ gives the source term, keeping the kernel term unchanged. The integrand receives a $\log|p-q|$ singularity feature and the frozen shape latent of our kernel. It reaches $6.7$ / $5.2$ (test mean / median), no better than the source lift (Table~\ref{tab:ladder}), while its volume quadrature takes $O(N_p N_q)$ network evaluations per problem; the boundary term, by contrast, is a single quadrature over a few hundred surface samples. Its boundary derivative $-\partial_\nu G_\theta$, computed on the $205$ Laplace test pairs by automatic differentiation, integrates to ${\sim}10^{-3}$ over the boundary (the Poisson kernel has mass $1$) and is negative on roughly half of it, so it satisfies neither defining property of the Poisson kernel (nonnegativity and unit mass).

\subsection{Origin of the kernel-fit residual}\label{app:residual-origin}

\rev{Why does the 2D kernel leave a residual that the lift must correct? The KDE targets on which the 2D kernel is trained, and its boundary nodes, are built on simplified polyline contours of the digits, which do not coincide with the rasterized masks on which the reference solutions are computed. Two measurements locate the resulting error in this choice of geometric representation rather than in the sampler. First, integrating the targets themselves as if they were the kernel reproduces the reference solutions only to $5.1\%$, unchanged with $100\times$ more walks, so the error is systematic rather than statistical. Second, walks run directly on the mask geometry match the reference solutions to $0.07\%$, so the WoS estimator, its $\varepsilon$-shell, and the step cap are not responsible. The kernel reaches this ceiling, so the residual $e_h$ is the systematic error of its training signal rather than something the kernel fails to fit.}

\rev{On Laplace pairs, $u - u_h$ equals $e_h$, so the solution-level supervision of the lift, and of the residual head $r$ (Appendix~\ref{app:ladder}), contains exactly this error, which is why the learned correction removes most of it ($71\%$ for the single lift). Placing the KDE's boundary nodes on the contour of the rasterized masks instead, with the construction otherwise unchanged, lowers the kernel-only error from $5.9\%$ to $3.2\%$ in a controlled 2D experiment, and a kernel-plus-lift model (without $r$) trained on this kernel reaches $1.8\%$ / $1.7\%$ (mean / median) on test and $2.1\%$ / $2.2\%$ on OOD, against $2.1\%$ / $2.0\%$ and $2.6\%$ / $2.5\%$ for the model of Table~\ref{tab:mnist_main}. We leave this choice of geometric representation for the training signal, together with exploiting the linearity of $e_h$ in the design of $r$, to future work.}
\end{revblock}

\end{document}